\documentclass[10pt,journal,compsoc]{IEEEtran}

\ifCLASSOPTIONcompsoc

  \usepackage[nocompress]{cite}
\else

  \usepackage{cite}
\fi

\ifCLASSINFOpdf
  \usepackage[pdftex]{graphicx}

\else

\fi

\usepackage{amsmath}
\usepackage{xcolor}
\usepackage{stfloats}
\usepackage{adjustbox}

\newtheorem{definition}{Definition}%

\begin{document}

\title{FICNN: A Framework for the Interpretation of Deep Convolutional Neural Networks}

\author{Hamed Behzadi-Khormouji \hspace{0.5cm} José Oramas 
\IEEEcompsocitemizethanks{\IEEEcompsocthanksitem University of Antwerp - imec, IDLab - Department of Computer Science, Sint-Pietersvliet 7, Antwerpen, 2000, Belgium.\protect\\
E-mails: \{Hamed.behzadikhormouji, jose.oramas\}@uantwerpen.be
}
}

\markboth{ This is the author's version of an article submitted to IEEE Transactions on Pattern Analysis and Machine Intelligence}%
{Shell \MakeLowercase{\textit{et al.}}: Bare Demo of IEEEtran.cls for Computer Society Journals}

\IEEEtitleabstractindextext{
\begin{abstract}
With the continue development of Convolutional Neural Networks (CNNs), there is a growing concern regarding representations that they encode internally. Analyzing these internal representations is referred to as \textit{model interpretation}. While the task of \textit{model explanation}, justifying the predictions of such models, has been studied extensively; the task of \textit{model interpretation} has received less attention. 
The aim of this paper is to propose a framework for the study of interpretation methods designed for CNN models trained from visual data. More specifically, we first specify the difference between the \textit{interpretation} and \textit{explanation} tasks which are often considered the same in the literature. Then, we define a set of six specific factors that can be used to characterize 
interpretation methods. Third, based on the previous factors, we propose a framework for the positioning of interpretation methods. 
Our framework highlights that just a very small amount of the suggested factors, and combinations thereof, have been actually studied. Consequently, leaving significant areas unexplored.
Following the proposed framework, 
we discuss existing interpretation methods and give some attention to the evaluation protocols followed to validate them. Finally, the paper highlights capabilities of the methods in producing feedback for enabling interpretation and proposes possible research problems arising from the framework.      

\end{abstract}

}

\maketitle

\IEEEdisplaynontitleabstractindextext
\IEEEpeerreviewmaketitle

\IEEEraisesectionheading{\section{Introduction}
\label{sec:introduction}}
In recent years, there have been an increasing interest towards understanding features internally encoded by Convolutional Neural Networks (CNNs) deployed in critical applications, e.g.  Covid-19 detection from X-Ray Images~\cite{medical_app_protopnet}, pedestrian detection~\cite{pedestrian_app_protpnet}, etc.

This task of analyzing the features encoded internally in a model has been referred to by the \textit{interpretation} and \textit{explanation} terms,  interchangeably~\cite{interpretation_survey,interpretation_survey_2,interpretation_survey_3}. 
While~\cite{interpretation_survey_2} and~\cite{interpretation_survey_3} indicate existing discordant definitions regarding  \textit{interpretation} and \textit{explanation} in the literature, these works do not elaborate on the differences between them. Moreover, these works follow the common practice of using these terms interchangeably.
\cite{interpretation_survey_2} suggests \textit{semi-interpretability} as a transition between \textit{local interpretability} methods, that take as input a single image and justify predictions from it (model explanation); and \textit{global interpretability} methods, which explain the network/model as a whole (model interpretation). In addition, it considers any feedback related to these tasks as an \textit{explanation}. In contrast, \cite{xai_survey,taxonomies_shortcoming} refer to the same tasks as \textit{local explanation} and \textit{global explanation}, respectively. 

\begin{figure}
    \centering
    \includegraphics[width=1\linewidth]{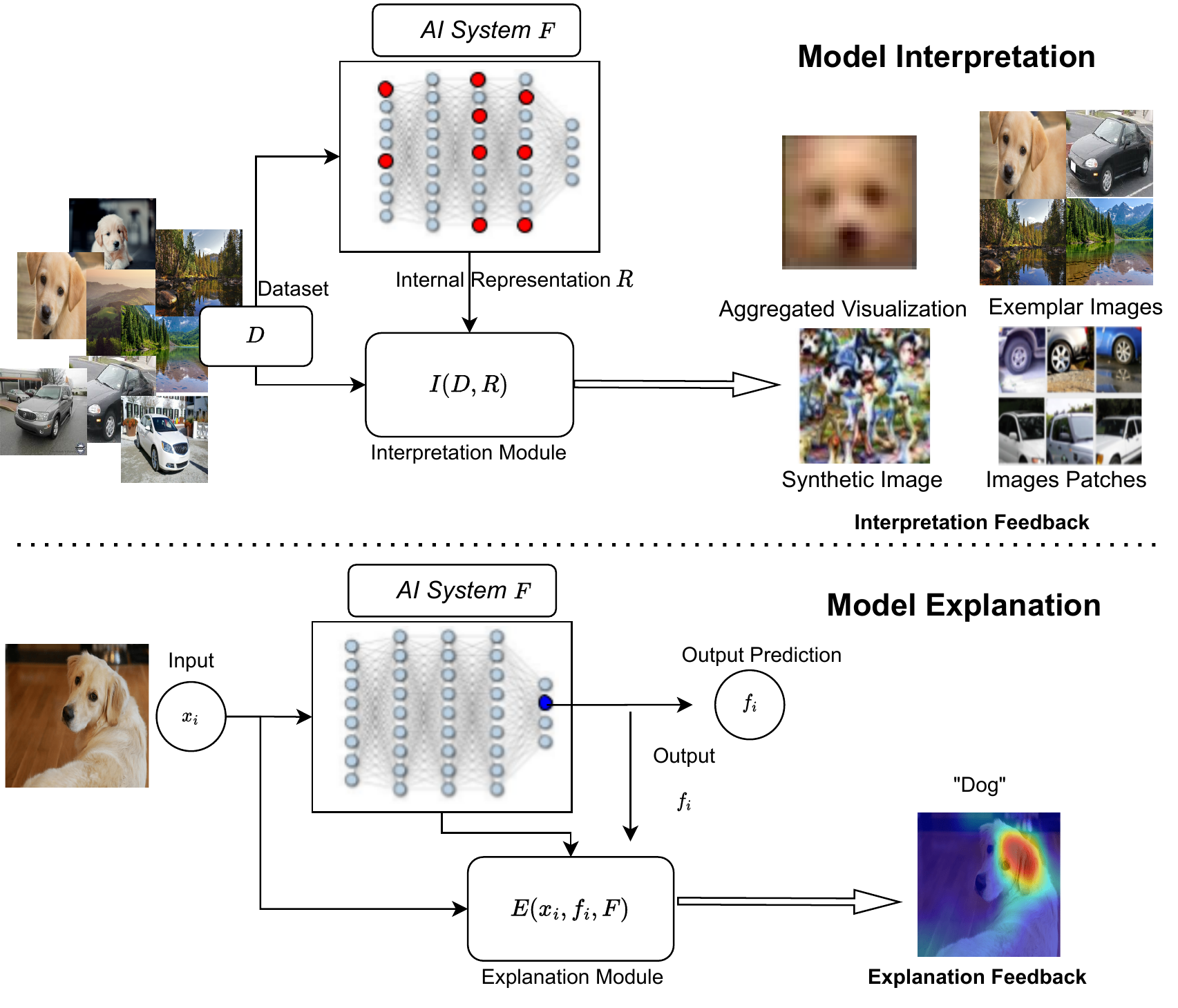}
    \caption{Model Interpretation (top) investigates internal encoded features being critical for model functioning. Model explanation (down) indicates characteristics of an input sample affecting model decisions.}
    \label{fig:teaser}
\end{figure}

As can be noted, despite surveying a common set of existing methods, these studies consider different perspectives to define  the explanation and interpretation tasks. As a result, there is no unified agreement on the exact definition of specific terms which lends itself to confusion. 
To address this weakness, we will begin by providing a specific definition for \textit{model explanation} and \textit{model interpretation} tasks as follows:

\begin{definition}[Model Interpretation]
Given a set of internal representation $R$ from an existing [pretrained] model $F$ and training data $D$, the main goal of the model interpretation task is to determine what a model has actually learned. That is, what informative characteristics or features from the training data $D$ that the model encodes into the internal representation $R$. In practice, this is related to producing insights into what internal relevant features within $R$ are critical for model functioning (\figurename~\ref{fig:teaser} top).
\end{definition}

\begin{definition}[Model Explanation]
Given a specific input example $x_i$, a model $F$, and an output prediction $f_i$ produced by the model $F$, the model explanation task aims at justifying why a given prediction/decision $f_i$ has been made for that input by the model. In practice, this relates to indicating what characteristics of the input example were taken into account by the model for making the produced prediction (\figurename~\ref{fig:teaser} down).
\end{definition}

To date, a considerable number of surveys/taxonomies have been put forward along the research line  of model explanation~\cite{expltaxonomy,gilpin2018explaining, expl_survey4, fmo,expl_meta_survey}. 
\cite{gilpin2018explaining} debates on the definition of explanation along with the evaluation protocols followed by methods from this research line.
This is complemented by the proposal of fundamental concepts that could assist their categorization. 
\cite{expl_meta_survey} conducts a meta survey on the existing model explanation surveys. More recently, \cite{taxonomies_shortcoming} discussed the shortcomings of existing taxonomies from this research line and proposed a new one.

In comparison to the model explanation research line, less attention has been paid to the model interpretation methods~\cite{interpretation_survey,interpretation_survey_2,interpretation_survey_3}.  There are also some studies about interpretation of Generative Adversarial Networks (GANs)~\cite{gan_lab,gan_dissection}. In this study, we focus on methods for visual interpretation of convolutional neural networks.  Despite the fact that these works have investigated some aspects related to the model interpretation task, they suffer from the following weaknesses.
The first weakness is related to the non-standard use of terminology which was mentioned earlier. As a second weakness, they encompass both model explanation and interpretation methods as interpretation methods. However, these two groups tasks have different characteristics and goals. 
Third, a side-effect introduced by the wide coverage of both explanation and interpretation methods is that these surveys need to rely on very coarse factors in order to be able to position the covered methods with respect to each other. For example,~\cite{interpretation_survey_2} defines \textit{local/global interpretability} factors indicating whether the method provides the output for a given individual sample or whole of the dataset. As another example,~\cite{interpretation_survey_2} and~\cite{interpretation_survey_3} define a factor indicating whether the method is applied during the training phase or after the model being trained. This factor is refereed by different terminologies namely \textit{passive/active methods}~\cite{interpretation_survey_2} and \textit{Specific/Agnostic models}~\cite{interpretation_survey_3}. 
Finally, as fourth weakness, existing surveys consider a small range of  interpretation methods. This constitutes a significant gap given the growing number of interpretation-related efforts occurring in recent years.

To address these weaknesses, we propose a framework for categorizing model interpretation methods. The framework introduces a set of six specific factors targeting the interpretation capabilities that these methods provide. Then, these factors are used as axes for the positioning of existing interpretation methods.
First, among these factors we consider \textit{feedback modality} as the means used by interpretation methods to provide feedback on the extracted insights, i.e. the relevant features. 
Second, we further analyze the level of  \textit{semanticity} of the provided feedback. This is a factor usually overlooked in existing surveys. 
Third, we cover a wider range of the interpretation methods, giving attention to the latest methods and diagnose active research lines.
Finally, we conduct a discussion within and across the groups defined by the proposed factors. In doing so, we uncover research gaps in each group as well as in the visual model interpretation research line as whole. Additionally, we suggest some potential solutions for addressing the identified gaps.

This paper is organized as follows: Section~\ref{sec:framework} introduces the framework. In this section, a set of six factors are defined. 
This is followed by the grouping of covered interpretation methods on the proposed factors and their detailed description. Section~\ref{sec:discussion} provides a discussion over the covered model interpretation methods respective to each defined factor. Furthermore, we touch on the  evaluation protocols followed for the validation of each method. Section~\ref{sec:related_work} provides an overview on the surveys in both, model explanation and interpretation, research lines which are used for positioning our work. Finally, the paper is concluded in Section~\ref{sec:concolusion}.

\section{A Framework for Model Interpretation Methods}
\label{sec:framework}

We begin this section by describing the factors that will serve as axes for the categorization of model interpretation methods.
Then, based on these factors we gradually categorize the methods.
At the end of this section, we summarize the factors of the discussed interpretation methods in Table~\ref{table:qualitative_properties}.

Our framework characterizes existing interpretation methods based on the following factors. 

\begin{itemize}

    \item \textbf{Interpretation Capability Integration.} This factor describes the point on which interpretation capabilities are added to a given base model. Two options are possible. On the one hand, interpretation capabilities can be provided after the base model has been trained, i.e. in a \textit{Post-Hoc} manner ~\cite{vebi}, \cite{networkdissection}, \cite{net2vec}, \cite{ace}, \cite{topicmodel}, \cite{pace}. On the other hand, specific mechanisms can be added to the base model at design time, i.e. prior training, so that the resulting model is interpretable post-training. Thus, producing a model that is \textit{interpretable-by-design} or \textit{inherently-interpretable}~\cite{casebasedreasoning}, \cite{protopnet}, \cite{conceptwhitening}, \cite{prtomil}, \cite{attribute_prototype}.

    \item \textbf{Task Specificity.} 
    This factor refers to whether the interpretation mechanism depends, i.e. \textit{task-specific}, or not, i.e. \textit{task-agnostic}, on characteristics related to the task addressed by the based model. 
    For the case of the  classification task this factor could indicate whether the interpretation mechanisms are dependent on each individual class of interest (\textit{class-specific}) ~\cite{vebi}, \cite{protopnet}, \cite{ace}, \cite{conceptwhitening}, \cite{pace}, \cite{prtomil} or whether it is general across the dataset (\textit{class-agnostic})~\cite{networkdissection}, \cite{casebasedreasoning}, \cite{net2vec}, \cite{topicmodel}.

    \item \textbf{Feedback Semanticity.} A \textit{concept} can be defined as an idea associated with properties that are \textit{semantically meaningful} for humans~\cite{conceptpaper}. In computer vision problems, this \textit{semantically meaningful} property can be presented in different forms such as annotation masks with text labels \cite{networkdissection}, bounding box with assigned captions \cite{coco}, or part-level annotations \cite{WelinderEtal2010}. This factor describes whether the feedback provided by a given interpretation method can be associated to a semantically meaningful concept~\cite{networkdissection}, \cite{net2vec},\cite{compositionalexplanations}. This also includes whether such meaningful semantics can be assigned/mapped to the internal units of the base model.

    \item \textbf{Annotation Dependance.} This factor describes the level of annotation required by the interpretation methods in order to operate. For the case of the image classification task~\cite{surveyimgeclfic}, some methods depend on image-level annotations~\cite{vebi},~\cite{subnetworks} originally used to train the base model, while others on additional detailed pixel-level annotations~\cite{networkdissection}, \cite{net2vec}, \cite{compositionalexplanations}.

    \item \textbf{Architecture Coverage.} 
    This factor indicates the level to which the architecture of the base model being interpreted is considered when analyzing the encoded representation. In this regard, interpretation methods may consider the whole architecture~\cite{vebi}, \cite{networkdissection}, \cite{net2vec} or focuses only on specific parts from it~\cite{casebasedreasoning}, \cite{protopnet}, \cite{ace}, \cite{topicmodel}, \cite{conceptwhitening}, \cite{pace}, \cite{prtomil}.

    \item \textbf{Feedback Modality.} This factor describes the modality being used by interpretation methods to provide feedback from the insight extracted from the information internally encoded in the base model. This modality can be a quantitative measurement such as contribution of identified relevant features on the model performance~\cite{linear_probing} or different forms of visualization. We refer to this as interpretation visualization. Examples of different visualizations of the interpretation feedback are synthetic images~\cite{feature_inversion,network_inversion}, average of visualizations~\cite{vebi}, extracted superpixel~\cite{ace}, image patches~\cite{topicmodel,pace}, heatmap visualization~\cite{protopnet,tesnet,deformable_protopnet}, or examples of input images~\cite{conceptwhitening}.
    
\end{itemize}

In what follows, using the proposed framework, we first divide existing efforts based on the  \textit{Interpretation Capability Integration} factor, i.e. as either \textit{Post-Hoc} or \textit{Interpretable-by-Design}. Then, the rest of the factors will be discussed within each of these two categories.

\subsection{Post-Hoc Interpretation Methods}

\subsubsection{Class-Specific}
\label{post-hoc_cs}
\textit{Class-Specific} methods provide interpretability, as the name suggests, at the level of classes. Put it differently, the methods identify features in the latent representation for each of the classes of interest. 
There is a group of post-hoc interpretation methods that apply the approach of internal representation inversion. These methods aim to generate synthetic images from the internal representations to show visually the features encoded by the models. These methods can be classified as  \textit{Class-Specific}~\cite{class_scoring_model}~\cite{non_parametric_patch_prior} or \textit{Class-Agnostic}~\cite{network_inversion}. The \textit{Class-Agnostic} methods will be discussed in Section~\ref{sec:post_hoc_Class_Agnostic}. 

An example of the internal representation inversion in \textit{Class-Specific} category is~\cite{class_scoring_model}. The method designs an image reconstruction loss function to reveal class-relevant features learned by a model. We refer to this method as \textit{Class Scoring Model}. The proposed approach tries iteratively to estimate a natural image from an initially randomized image such that the output score of the given class for that image is maximized. The resulting image depicts content relevant to a target class learned by the model.

~\cite{non_parametric_patch_prior}  extended the Class Scoring Model~\cite{class_scoring_model} and Feature Inversion~\cite{feature_inversion} (discussed in Sec. \ref{sec:post_hoc_Class_Agnostic}) by adding a non-parametric patch prior to their regularization term to improve the reconstructed images. Also, they consider the activations from fully connected layers, while the Feature Inversion and the Class Scoring Model utilize the activations from convolutional filters and output logits in their optimization procedure, respectively. As output, the Class Scoring Model~\cite{class_scoring_model} and \cite{non_parametric_patch_prior} generate visualizations of internal representations. 
The visualizations reveal some patterns similar to those present in the dataset seen by the model, which are understandable by humans. However, the visualizations produced by these methods suffer from noise and unclear patterns that lend themselves to confirmation bias and introduce subjectivity.
 
Another example of this category is \cite{mid_level_vp}. It conducts association rule mining via the \textit{Apriori} algorithm~\cite{apriori}, as a means to identify frequent visual patterns encoded within a model. To do so, it utilizes the activations computed by a fully-connected layer from cropped image patches fed to the model. These patches are then grouped into two categories: the target class and the background (including patches from other classes) which is followed by the creation of a binary transaction database from them. Each transaction contains the indices of neurons in the fully-connected layer, along with an extra item indicating the index of one of the two categories (binary transaction database). Finally, visual patterns are identified by extracting  frequent itemsets of the indices in the fully-connected layer from the transaction dataset.

Aiming to identify internal relevant features of each class of interest, VEBI~\cite{vebi}, opposite to \cite{mid_level_vp}, utilizes the feature maps produced by all layers of a CNN model.
VEBI enables model interpretation through two steps: 1) class-specific relevant feature identification, and 2) visual feedback generation. The identification of relevant features is formulated as a $\mu$-LASSO problem where indicators $\omega_c$ are obtained for the aggregated internal activations produced by each image $x_i$ from a given class $c$. Visual feedback of these relevant features is produced via average visualizations produced from image crops extracted from regions where the identified relevant features have high activation.

In contrast to \cite{vebi}, TCAV~\cite{tcav} proposes \textit{Concept Activation Vectors (CAV)} that enables model interpretation with a partial coverage of the architecture that defines the base model. 
To compute CAVs, the dataset is re-grouped to define a binary setting in which one group is composed by images illustrating visual patterns of interest in one class (target class), and the other by a set of random images as other class. 
Then, a linear classifier is trained to classify the activations of these images as computed by the neurons of a given layer. The resulting classifier, called CAV, is a vector with the same length as the number of neurons in the layer serving as input. This vector highlights the representation of the class of interest in the considered layer. Then, the method uses the CAVs and directional derivatives to calculate the sensitivity of the target class to the CAV. To do so, it computes the difference between output logits of the target class for the original activations and the modified activations, i.e., the summation of the CAV activations and the original activations. This quantitatively measures the sensitivity of the model with respect to the representation of any class in the given layer of interest. Finally, visual feedback is provided via examples of the target class whose activations of the target convolutional layer has higher similarity to the obtained CAV. 

In contrast to VEBI~\cite{vebi} and similar to TCAV~\cite{tcav}, ACE~\cite{ace} enables model interpretation with a partial coverage of the architecture that defines the base model. It utilizes the k-means algorithm to cluster the feature maps of a given layer for a subset of image superpixels belonging to a given class. The superpixels corresponding to each cluster refer to similar visual patterns depicted in the input images.

Instead of applying a clustering approach as in ACE,  Invertible Concept-based Explanation (ICE)~\cite{ice} extends ACE by decomposing the feature maps computed from the last convolutional layer via Non-negative Matrix Factorization (NMF)~\cite{nmf}.
Following NMF, the feature maps $A{\in}R^{h \times w \times d}$ are decomposed into three components namely dimension-reduced feature maps $A'{\in}\Re^{h \times w \times d'}$, dimension reducer matrix $V{\in}\Re^{d' \times d}$ and the residual error $U$. During the training phase, the reducer matrix $V$ is trained on the images from a given class in order to decrease the complexity of the feature maps' channels $d$. 
Afterwards, the parameters of the reducer matrix are fixed and considered as class-relevant vector bases representing directions for different representations in the latent space. At  test time, given computed feature maps for a set of test images, the reducer matrix is applied to generate new feature maps with lower number of channels (i.e., $A'{\in}\Re^{h \times w \times d'}$) for each feature maps pertaining to an image. Then, Global Average Pooling is applied on each channel of the new feature map and the resulting value is considered as its score. The images with higher scores are selected. Then, for each selected image, the channel with the highest score is chosen and binarized. The binarized map is resized to the size of the image and overlaid over it 
to illustrate the extracted visual pattern. The fidelity of the learned reducer matrix is evaluated by measuring the effect that using inverted feature maps, computed via NMF, have on classification accuracy. In addition, the consistency of the patterns depicted in the generated visualizations is evaluated in a user study.

Concept Attribution~\cite{concept_attribution} identifies class-specific internal units in two stages. In the first stage, it learns a global input feature occluder for a given class which changes the prediction on the image with the lowest input feature perturbation. In the second stage, it aims to assign a class-specific weight to each convolutional filter via the obtained global input feature occluder. To do so, it aggregates the difference between activations computed from the original input images and the modified (occluded) ones. These differences are then considered as the weights of their respective  convolutional filters. To enable visualization of the filters with the highest score, it utilizes the technique of internal representation visualization~\cite{inceptionism} to synthesize images which maximise the activations generated by these filters. These synthetic images illustrate the features encoded by the identified filter.  

\cite{subnetworks} provides interpretation by extracting class-specific subnetworks which include critical internal units relevant to a given target class. Here we refer to this method as \textit{Critical Subnetworks}. To do so, the method assigns  gates/weights to each internal convolutional filter in the base model. These gates are expected to represent importance weights of those filters for a given class. 
During the training phase of these gates, the output feature maps are multiplied by their corresponding gate resulting in new feature maps which are passed to the higher layers. 
The method learns these gates by minimizing binary cross-entropy (BCE) loss such that the extracted subnetwork has an accuracy close to that of original network while it deactivating class-irrelevant filters. In addition, to extract each subnetwork for a given class, a binary dataset is created including images from the target class (i.e., positive class) and images from the rest of the classes (i.e., negative class).

\cite{pace} has proposed a framework, called PACE, which is defined by a set of autoencoders whose latent representation vectors are trained specifically with respect to each class of interest. The encoder components transform the feature maps of the input images computed by a part of the model (i.e., one convolutional layer), into the latent space vectors. Then, the decoder components project the vectors back to the space of the convolutional feature
map. The learning of relevant features per class occurs
in these latent space vectors. This is achieved by measuring the similarity matrices between representations of the latent space vectors and the encoder representations. The similarity matrices w.r.t the learned representations of a given class can be treated as
explanation masks to recognize the relevant region in the input images after suitable resizing.

\textbf{Discussion.} While VEBI~\cite{vebi} and PACE~\cite{pace} utilize the images of all classes to identify/learn relevant features in the latent representations, the Class Scoring Model~\cite{class_scoring_model},~\cite{non_parametric_patch_prior},~\cite{mid_level_vp}, TCAV~\cite{tcav}, ACE~\cite{ace}, Critical Subnetworks~\cite{subnetworks},  ICE~\cite{ice}, and Concept Attribution~\cite{concept_attribution} require to be run in separate stages considering image examples from one class at a time. 

In addition, regarding the \textit{Annotation Dependency} factor, the  Class Scoring Model~\cite{class_scoring_model},~\cite{mid_level_vp}, VEBI~\cite{vebi}, TCAV~\cite{tcav}, Critical Subnetworks~\cite{subnetworks} and PACE~\cite{pace} rely only on image-level labels. In contrast, \cite{non_parametric_patch_prior}, ACE~\cite{ace}, ICE~\cite{ice}, and Concept Attribution~\cite{concept_attribution} are completely independent of any image label and/or pre-defined annotations. 

Furthermore, regarding the \textit{Feedback Modality} factor, the methods utilize different modalities to enable interpretation of the internally-encoded representations. %
For example, generating synthetic images depicting the internal representations in the Class Scoring Model~\cite{class_scoring_model} and \cite{non_parametric_patch_prior}, creating average visualizations of image crops w.r.t an internal unit in VEBI~\cite{vebi}.
ICE~\cite{ice} highlights regions of input images using binarized heatmaps. Other works provide visualizations in the form of similar visual patterns in the dataset (i.e. exemplars) such as image patches in \cite{mid_level_vp}, PACE~\cite{pace}, clusters of superpixels in ACE~\cite{ace}, synthesised images maximizing the activations in TCAV~\cite{tcav} and Concept Attribution~\cite{concept_attribution}. Also, providing examples of the dataset in TCAV~\cite{tcav} is another modality to provide interpretation feedback. Examples of the feedback modality of each method can be seen in \figurename~\ref{fig:si}, \ref{fig:ip_2}-\ref{fig:aic_hv}.  

Regarding the \textit{Feedback Semanticity},
none of the methods guarantees that the provided interpretation feedback will have a semantic meaning.

Regarding the \textit{Architecture Coverage} factor, while VEBI~\cite{vebi}, Concept Attribution~\cite{concept_attribution}, and Critical Subnetworks~\cite{subnetworks} enable interpretation by considering the feature maps of all the convolutional layers, the other reviewed works only consider the feature maps of a small part of the model, thus reducing the level of interpretation of the model. Furthermore, with the exception of~\cite{non_parametric_patch_prior},~\cite{mid_level_vp}, VEBI~\cite{vebi}, and Critical Subnetworks~\cite{subnetworks}, none of the discussed works are able to link
the internal units of the base model with the identified/learned relevant representations, thus reducing their intelligibility.

\subsubsection{Class-Agnostic}
\label{sec:post_hoc_Class_Agnostic}
In contrast to \textit{Class-Specific} methods, \textit{Class-Agnostic} methods enable interpretation by identifying relevant latent elements without exploiting or imposing class-specific constraints. 

Early works of this category followed the internal representation inversion approach, more specifically Feature inversion~\cite{feature_inversion} and Network Inversion~\cite{network_inversion}.  \cite{feature_inversion} put forward the \textit{feature inversion} approach to reconstruct an image from the internal representation of a given convolutional filter. The resulting image aims to reveal features learned by a convolutional filter. To do so, the method applies an internal representation reconstruction loss function which considers the Euclidean distance between internal representations of a given image and the reconstructed one plus a regularizer enforcing a natural image prior. This loss function is minimized using the gradient descent technique.

Different from Feature Inversion~\cite{feature_inversion} which minimizes an internal representation reconstruction error for a specific target image in the dataset, the Network Inversion method~\cite{network_inversion} minimizes an image reconstruction error. To do so, the loss function in Network Inversion measures the intensity difference between an original image and the reconstructed one by a network. The network is fed by the internal representation of a given convolutional layer and outputs a reconstructed image which is the input to the defined loss function. In this method, the  network weights are trained in such a way that the loss error between the original images and their reconstructed counterparts is minimized.

The methods explained above namely; Feature Inversion~\cite{feature_inversion} and Network Inversion~\cite{network_inversion} have studied the encoded features by generating visualizations from inverted internal representations. There are other \textit{Class-Agnostic Post-Hoc} methods proposing different methodologies to interpret the encoded features. 
For instance, even thought it is not a model interpretation method per se, the methodology used by \cite{object_detectors} for investigating the emergence of object detectors in models is quite related.
More concretely, given a convolutional filter, the method, first, collects the images that are highly activated by that convolutional filter. 
Second, given each of the images in the set, the method randomly selects multiple patches of the images and pushes them through the model. 
Third, the difference between activations produced by the original image and its patches is computed and considered as discrepancy map. Next, the average of all discrepancy maps are calculated. 
Finally, a region of interest is highlighted by the average map for all the collected images. 
Also, in this work, convolutional filters are annotated by a semantic label in a user-based procedure.

Taking the analysis above a bit deeper, \cite{visual_att_indicator} identifies sparse sets of internal units encoding visual attributes. To do so, it utilizes activation maps of all the images in the dataset produced by all the layers of a CNN model. Then, it formulates a $\mu$-LASSO problem to learn an indicator $\omega_j$ for each annotated visual attribute $j$. The indicator $\omega_j$ represents the indices of internal units encoding the visual attribute $j$. 
Being the method that served as inspiration to VEBI~\cite{vebi}, led to both methods having some similarities. 
However, the main difference between \cite{visual_att_indicator} and VEBI, lies in the fact that the former is independent of class labels related to the original classification task. Therefore, it is capable of identifying class-agnostic units which represent visual attributes.

Linear Probes~\cite{linear_probing} investigates the dynamics of convolutional layers. Feature maps are extracted from each convolutional layer, and a linear classifier is fitted to each layer separately in order to determine the class label. This is done in order to investigate the linear separability of each layer. By studying the accuracy of each classifier, it was observed that the degree of linear separability of the feature maps of the layers increases as we reach the deeper layers.

Different from the Linear Probes that map an internal representation to a class label, \cite{revers_probing} proposed a Reversed Linear Probing approach that maps concept labels into a quantized internal representation. 
To do so, first, the method utilizes the \textit{k-means} algorithm to cluster the internal representations of a given convolutional layer, pertained to the input images, in order to quantize the representations into cluster indices. Second, in order to introduce the concepts in the form of discrete attribute vectors (i.e., presence or absence of a concept), the method utilizes a set of pre-trained models. The output prediction, the predicted classes, of all the pre-trained models for a given input image are concatenated to form an attribute vector.
As a result, there is an attribute vector, as concept label, and a cluster indices vector for each image and its internal representation. Finally, the method trains a linear model to map attribute vectors, as concepts, to cluster indices as the quantized form of the internal representations for the given images. Consequently, it can analyze the concepts encoded in the internal representation of a given input image. Hence, an image is linked with the concepts vector predicted by the linear model for its corresponding quantized representation. 

Both Linear Probing and Reversed Linear Probing provide interpretation in the form of a set of classifiers to understand the internal representations. However, they are not able to assign a human-understandable concept to the internal units, thus reducing the level to which they enable interpretation of the base models.

\cite{selectivity_index} enables interpretation by assigning a quantitative factor, called Selectivity Index, to each convolutional filter in a given pre-trained model. The Selectivity Index is defined based on the activations produced by a convolutional filter for a pre-defined number of  cropped images whose activations are higher than other cropped images in the dataset. The patches of the images used for each convolutional filter covers a variety of class labels. Hence, the Selectivity Index can be used to quantitatively analyze each convolutional filter either in a \textit{Class-Specific} or \textit{Class-Agnostic} manner. This is done by calculating the relative frequency of each class via activations of its image patches in the set. To measure this frequency, first the relative activation of the image patches w.r.t a filter is obtained. It is defined by the fraction of the activations of the patch from the maximum activation computed by the filter from all the patches in the set.  Second, a normalized summation of the relative activations of the image patches for each class is computed  individually. Finally, the number of classes whose summation of relative frequency is higher than a threshold is selected as class label(s) for the investigated filter. In this line, each internal unit can be linked to either a specific class label or multiple class labels. The average of the extracted patches of the images for each unit is then calculated to show the patterns learned by that convolutional filter. 

Different from the above mentioned methods, there is a group of the \textit{Class-Agnostic Post-hoc} methods that measure the alignment between internal representations and annotation masks. One example of this groups is Network Dissection~\cite{networkdissection}. This method is usually applied in conjunction with the Broden dataset which is composed by images from several existing datasets. This provides Broden with a vast variety of semantic concepts and their corresponding annotation masks for texture, objects, and object-parts. Using this external dataset, the dissection method measures the alignment between thresholded feature maps computed in the convolutional filters of a given base model with annotation masks from Broden. Then, the semantic label whose annotation masks have the highest overlap 
with the feature maps is assigned to the filter that produced the activations. As a result, a list is produced indicating the classes/semantic concepts from the Broden dataset that were matched by the internal activations of a given base model.

Inspired by Network Dissection, \cite{net2vec} proposed Net2Vec, a method for quantifying and interpreting the level to which meaningful semantic concepts are encoded by filters in CNNs. 
Different from Network Dissection which aims to link a single unit with an annotated semantic concept, in Net2Vec this assignment is done in a many-to-one manner. More specifically, the feature map $M_k(x_i)$ produced by the weighed sum of multiple $k$ filters are linked to a single concept. 
Compositional Explanations~\cite{compositionalexplanations} extended Network Dissection to find logical compositions of abstract concepts encoded by the convolutional filters of the last convolutional layer. The intuition is that a convolutional filter may not be just a detector for one concept, but rather a composition of complex detectors characterizing multiple concepts. This is also different from Net2Vec which finds a combination of convolutional filters encoding only one concept. Compositional Explanations modified the Intersection-over-Union step, used by Network Dissection to measure overlap w.r.t. semantic concepts,  to consider logical composition operations such as disjunction (\textit{Or}), conjunction (\textit{And}), and negation (\textit{Not}) across different concepts. 
The resulting method is different from Network Dissection where only logical conjunction (\textit{And}) between an internal representation and one concept is considered.
Therefore, Network Dissection assigns each internal unit to a single concept, while Compositional Explanations assigns a convolutional filter to a composition of pre-defined concepts. Since, the problem of finding the best logical composition of concepts requires an exhaustive search, the method utilizes the Beam search algorithm to find an approximated solution to the problem.  

\cite{topicmodel} proposed an interpretation method, which is inspired by topic modeling methods~\cite{related2TopicModel_1}\cite{related2TopicModel_2}. Thus, throughout this paper, we refer to it as \textit{Topic-based interpretation}. The method from \cite{topicmodel} discovers topics within the activations of a given layer which are shared across all the classes of interest. These topics at the same time represent visual patterns in the dataset. However, the visual patterns covered by the identified topics do not necessarily possess a semantic meaning. Thus reducing their interpretation capability.

Similar to PACE~\cite{pace}, from the \textit{Post-Hoc Class-Specific} category, \cite{intervention} utilizes  generative models to provide a level of interpretation for the encoded features. More specifically, it applies a discrete variational autoencoder (DVAE) on the feature maps of a given layer to learn a binary compressed representation that drives the predictions made by the base model. 
Given the binary nature of the compressed representation, the method applies an intervention mechanism such as a flip on the encoder output to modify the reconstructed image. Then, the originally reconstructed image is qualitatively compared with the modified one to detect whether there is any bias in the representations internally learned by the model. While this method is capable of generating visualizations from learned compressed representations, these visualizations do not necessarily possess a semantic meaning.

\textbf{Discussion.} Regarding the \textit{Annotation Dependency} factor, \cite{visual_att_indicator} depends on annotations of visual attributes, 
while the attributes used in Reversed Linear Probing~\cite{revers_probing} are in the binary vectors  created by concatenating class labels predicted from a set of models. Network Dissection~\cite{network_inversion}, Net2Vec~\cite{net2vec}, and Compositional Explanations~\cite{compositionalexplanations} require expensive pixel-level annotations for their respective procedures. This dependency enables these methods to link internal units with a semantic meaning. In contrast, Linear Probes~\cite{linear_probing}, Selectivity Index~\cite{selectivity_index}, Topic-based interpretation~\cite{topicmodel}, and \cite{intervention} utilize image-level class labels in their interpretation procedure.

Regarding the \textit{Feedback Modality} factor utilized in the methods, Feature Inversion~\cite{feature_inversion} and Network Inversion~\cite{network_inversion} are able to generate a synthetic image for each given input image. This is similar to those of Class Scoring Model~\cite{class_scoring_model} and~\cite{non_parametric_patch_prior}, which can illustrate the patterns learned by internal units. 
Unlike these methods, Network Dissection~\cite{network_inversion}, Net2Vec~\cite{net2vec}, and Compositional Explanation~\cite{compositionalexplanations} highlight the regions for a set of input images whose activations, computed by the given convolutional filter(s), have the highest overlap with the annotations of a given semantic concept. 
In addition,~\cite{object_detectors} and Topic-based interpretation~\cite{topicmodel} produce visualizations of the image patches corresponding to the learned relevant features. Selectivity Index~\cite{selectivity_index} generates visualizations from the average of images patches corresponding to the identified internal relevant units. 
On the contrary, instead of producing image patch visualizations, \cite{visual_att_indicator}, Reversed Linear Probing~\cite{revers_probing}, and \cite{intervention} present a set of image exemplars containing similar patterns respective to identified internal units, learned clusters and decoders, respectively.
Using a different modality, Linear Probes~\cite{linear_probing} reports the accuracy of the learned classifiers as the feedback of analyzing the linear separability of the internal representations encoded in the layers. 
With the exception of Network Dissection~\cite{networkdissection}, Net2Vec~\cite{net2vec}, and Compositional Explanation~\cite{compositionalexplanations}, none of the above mentioned methods guarantee or provide a quantitative procedure that the highlighted/extracted regions in the image data necessarily possess a semantic meaning. Examples of the feedback modality of each method can be seen in \figurename~\ref{fig:si}-\ref{fig:aic_hv}.

Taking the capability of providing semantic feedback (i.e., \textit{Feedback Semanticity} factor) into account, it was observed that~\cite{object_detectors},~\cite{visual_att_indicator}, Network Dissection~\cite{networkdissection}, Net2Vec~\cite{net2vec}, and Compositional Explanations~\cite{compositionalexplanations} effectively address the task of linking internal units of a base model with semantic meaning.  This capability was not possible in the methods from Sec.~\ref{post-hoc_cs}. Moreover, excepting these works, none of other reviewed works in the \textit{Post-Hoc Class-Agnostic} category are able to associate semantic labels to the internal units of a given pre-trained model, nor assigning a semantic labels to the provided feedback, i.e., the generated visualization feedback of the interpretation.

Finally, regarding \textit{Architecture Coverage} factor, \cite{visual_att_indicator} considers feature maps produced from all convolutional layers to provide interpretation. Along this line, \cite{object_detectors}, Network Dissection~\cite{networkdissection}, and Net2Vec~\cite{net2vec}, although, provide insights for each convolutional filter and layer, respectively, they should be run separately on one or small sets of the filters at each time. The possible reason is that their methodology is able to provide interpretation for limited parts of a model instead of considering their relation. In contrast, other reviewed works in this category cover partially the representations in the architectures. Hence, these methods are not able to provide a complete interpretation of the base models.

\subsection{Interpretable-by-Design Methods}
\label{sec:intr_by_dsn}
Methods belonging to this category, follow the idea of designing learning algorithms so that the resulting model after training has specific properties that make it interpretable/explainable without the need of additional [post-hoc] processes.
Similar as before, we group the discussion of methods from this type based on the \textit{Task Specitivity} factor. Then, additional factors are discussed gradually within each group.

\subsubsection{Class-Specific}

One of the early deep \textit{Class-Specific Interpretable-by-Design} methods was proposed by \cite{capsul_net} with their proposal of Capsule Networks. 
Each layer contains a group of neurons called capsule.  The activity vector of these capsules encodes spatial information as well as the probability of an object or object-part being present. This is done by introducing an iterative routing-by-agreement mechanism which preserves the spatial relations among encoded features. In this mechanism, lower-level capsules model lower-level features coming from the input and link its output to capsules in the following layers. In order to determine which higher capsule should be routed to, the method defines a weight matrix representing the agreement relation between the lower capsules and the higher ones. Therefore, the lower capsule computes a prediction vector of the higher capsule by multiplying its own output by the weight matrix. 
Although, in theory, the method is expected to preserve the spatial relations among the encoded features, it does not provide any feedback modality to illustrate the learned spatial relations, nor measuring their alignment with semantic concepts.

Different from Capsule Networks, which introduce a new module/component (capsule layers) for inducing model interpretability, \cite{interpretable_conv_filter} proposes Interpretable Convolutional Filters. This method adds a new term to the training loss, called filter loss, to regularize the representations learned by each convolutional filter in a given layer. This filter loss pushes a convolutional filter to focus its activation on a specific spatial location of a class. To do so, it defines a set of masks with the same spatial resolutions of the feature maps computed by a given filter. Each mask follows a Gaussian distribution in a specific location as the ideal distribution of activations computed by that filter. Then, the filter loss is defined as the negative mutual information between a set of feature maps computed by the filter and the masks of the target class. Minimizing this loss guarantees that the  convolutional filter encodes a distinct  "part" of a class. Moreover, it constrains the activations of a filter to a single part of an object, rather than repetitively appear on different regions covering the object.

Taking the Interpretable Convolutional Filter~\cite{interpretable_conv_filter} into account, \cite{interpretable_decision_tree} enables interpretation of a CNN through the construction of class-specific decision trees along with the modification of the last convolutional layer to focus on the object parts. The decision tree is constructed based on a bottom-up hierarchical clustering approach. Here, we refer to this method as \textit{Interpretable CNN Decision Tree}.
In order to simplify the construction of the tree, the method re-trains the last convolutional layer using the filter loss proposed in \cite{interpretable_conv_filter} prior to following the procedure for constructing the tree. This step fine-tunes the last convolutional layer to recognize specific object parts. Next, the receptive fields of each convolutional filter are computed. Then the part from the images which frequently appears in the receptive filed is considered as the part label for the filter.

Regarding the construction of the tree, in the first step, each node encodes the specific rationale for the prediction for each image. 
In the second step, in an iterative manner, two nodes with the highest similarity are merged to create a new node in the second level of the tree. Then, for each new node, a linear loss function is defined to learn a sparse weight vector representing the convolutional filters. 
The resulting sparse vector shows the contribution of the filters to the predictions at a specific level of the tree for the set of images merged in the newly created node. %
Therefore, each node of the tree encodes the internal representations of the convolutional layer into elementary concepts of object parts. The method additionally measures the contribution of each of the parts to the prediction at each level of the tree using the learned sparse vector. 
Hence, each node in each level of the tree is considered as a partial description of the 
decision mode of the CNN on that level of the tree for a given prediction.
Finally, given a testing image the CNN computes an output score for the prediction. Then, the decision tree estimates the decision mode of the prediction at each level. Since each node in the tree has encoded a set of convolutional filters such that each one represents a specific part, thus each estimated decision mode explains the contribution value of the parts, presented at that level, to the prediction. In addition, it is able to highlight the image examples as well as image patches encoded in each node via visualizations of the receptive field whose convolutional filter indices have been indexed in that node. 

Different from Interpretable Convolutional Filter~\cite{interpretable_conv_filter} and Interpretable CNN Decision Tree~\cite{interpretable_decision_tree} which use a loss term to steer the last convolutional layer to learn interpretable representations, \cite{protopnet} has introduced a network architecture which includes a new interpretable module. The architecture called prototypical-part network (ProtoPNet), includes a prototype layer between the last convolutional layer and the fully connected layers. 
This layer includes class-specific trainable vectors which are responsible for learning prototypical parts of their target class. 
  
During the training phase, given an input image, the model is able to identify several parts of the input 
with high similarity with 
trainable prototypical vectors of some classes. To do so, the method follows an iterative training algorithm composed by two stages. In the first stage, keeping the classifier parameters fixed, the method optimizes jointly the parameters of the convolutional filters and the prototypical vectors in the prototype layer. The proposed loss function computes $L^2$ distance between the feature maps of the last convolutional layer and each prototypical vector in order to cluster the representation of the images around the similar prototypical parts of the their ground-truth classes. In the second stage, keeping the parameters of the convolutional filters and the prototype layer fixed, the method optimizes the parameters of the classifier in order to classify the input images via the learned prototypical vectors. 
Moreover, the method can generate a heatmap visualization of the learned prototypical parts for each input image.

Instead of learning prototypical parts of the input images in ProtoPNet~\cite{protopnet}, Concept Whitening~\cite{conceptwhitening} put forward a built-in module that is composed by whitening and orthogonal transformations. These operations aim at aligning the latent space of the internal units 
with similar visual patterns emerging in a predefined set of images. In this method, two types of data are considered for the training phase. First, $D=\{x_i,y_i\}_{i=1}^n$ is a dataset that includes $n$ samples and their labels for training the based model. Second, the $m$ auxiliary datasets $D_{1}, D_{2}, D_{3},...,D_{m}(D_{m}\subset D)$ where each one contains instances that depict a common visual pattern for optimizing the orthogonal transformation matrix. 
Similar to the ProtoPNet training algorithm~\cite{protopnet}, Concept Whitening has a training algorithm including two stages. In the first stage, the parameters of the base model are optimized. In the second stage, keeping the base model's parameters fixed, the parameters of the transformation module are optimized to separate the latent spaces of the last convolutional layer such that each direction in the latent space encodes a specific visual part in the dataset.   

Inspired from ProtoPNet, \cite{prtomil} combined the prototype layer structure of ProtoPNet and Attention-based MIL pooling. Their method, known as ProtoMIL, is capable of learning representations for a bag of instances. 

\cite{tesnet} have proposed TesNet architecture to improve the diversity and  discriminative properties of the prototype layer in ProtoPNet. To do so, first, an orthonormal constrain is added to the loss function to push prototypical vectors, within prototype layer, from different classes apart from each other. Second, the method applies Projection Metric~\cite{projection_metric}, a distance metric on the Grassmann manifold, to separate the prototypes of each pair of classes. These two constraints help to minimize the correlation between prototypes within and between each pair of classes. This method utilizes as same two-stage training algorithm as that of ProtoPNet~\cite{protopnet}.

Different from existing case-based and prototype-based methods namely; Case-based reasoning~\cite{casebasedreasoning}, ProtoPNet~\cite{protopnet}, Proto Tree~\cite{prtotree}, and TesNet~\cite{tesnet}, which use spatially rigid prototypes, Deformable ProtoPNet~\cite{deformable_protopnet} have recently proposed the use of spatially flexible prototypes. 
This property enables prototypical vectors withing prototype layers to adaptively change their relative spatial positions w.r.t to the input images. As a result, the prototypical vectors will be robust to variations in pose and context, i.e., detect object features with a higher tolerance to spatial transformations, as well as improve the richness of their visualizations. 
This spatial flexibility property is defined by an offset function which adds some offset to the location of input patches of feature maps, thus enabling each prototypical part to move around when it is applied on a spatial location. Moreover, following a similiar orthogonality loss as in TesNet~\cite{tesnet}, Deformable ProtoPNet defines an orthogonality loss between all the  prototypical vectors within a class to avoid overlapping between them. This is different from TesNet which applies this property among each pair of prototypical vectors in a class-agnostic manner.

\textbf{Discussion.} Regarding the \textit{Annotation Dependency} factor, Capsule Networks~\cite{capsul_net}, Interpretable Convolutional Filter~\cite{interpretable_conv_filter}, Interpretable CNN Decision Tree~\cite{interpretable_decision_tree}, ProtoPNet~\cite{protopnet} and its \textit{Class-Specific} extensions namely, ProtoMIL~\cite{prtomil}, Tesnet~\cite{tesnet}, and Deformable ProtoPNet~\cite{deformable_protopnet} as well as its \textit{Class-Agnostic} extension ProtoTree~\cite{prtotree} (reviewed in Sec.~\ref{sec:int_b_dsg_ca}) depend only on image-level labels. 
In contrast, CW~\cite{conceptwhitening} requires, in addition to the image label, an extra dataset for the procedure of latent space alignment, i.e., training the transformation module. 

Furthermore, it should be noted that ProtoPNet~\cite{protopnet} and the extensions mentioned above apply a built-in module after the convolutional layers to provide flexibility in the feature learning process taking place in those layers. In a similar, in the context of capsule networks, specific components are present between capsule layers that determine the way activations are routed across layers. In contrast, CW~\cite{conceptwhitening} enforces each convolutional filter of a given layer to align its latent representation to specific input images. Complementary to the previous efforts, Interpretable Convolutional Filter~\cite{interpretable_conv_filter} defines a loss term to regularize the representations of convolutional filters, instead of injecting a direct transformation module. 

Regarding the \textit{Feedback Modality}, the output of each prototypical vector in the ProtoPNet~\cite{protopnet}, ProtoMIL~\cite{prtomil}, TesNet~\cite{tesnet}, and Deformable ProtoPNet~\cite{deformable_protopnet} can be visualized for the input images which have the closest patch to those encoded by the prototypical vectors. 
This visualization can be generated in two forms: image patch and heatmap visualization.  The heatmap visualization is generated by superimposing the output of the similarity map computed between a feature map and a prototypical vector on the input image. In addition, these methods highlight the patches of input images whose feature maps have the closest distance (highest similarity) to one of the learned prototypical vectors. \figurename~\ref{fig:ip_2}-\ref{fig:aic_hv} show examples of the feedback modality in each of the discussed method in this category.

In contrast, CW~\cite{conceptwhitening} provides only exemplar images, illustrating similar patterns, which produce the highest activation for a given convolutional filter. Moreover, Interpretable Convolutional Filter~\cite{interpretable_conv_filter} and  Interpretable CNN Decision Tree~\cite{interpretable_decision_tree} compute the receptive fields from a spatial locations  in the feature maps with the highest activation value from the filters  
 to highlight the encoded visual parts in the filters. This is done in order to show how each convolutional filter have been aligned to a specific visual pattern. Interpretable CNN Decision Tree~\cite{interpretable_decision_tree}, additionally, illustrate the examples of the dataset showing similar patterns. 

Different from the above, Capsule Networks~\cite{capsul_net} do not provide any feedback from learned representations in the capsule layers. A common practice to produce visualizations of the representation is by attaching a decoder and reconstructing the image when a given part of the representation is ablated.%

Regarding the \textit{Feedback Semanticity}, prototype-based methods only generate visualization of the learned prototypes. 
With the exception of Interpretable Convolutional Filter~\cite{interpretable_conv_filter} that takes advantage of Network Dissection~\cite{networkdissection} to quantitatively evaluate the semantic meanings of filters representations, 
there is no guarantee that the extracted patterns align w.r.t semantic concepts.

Finally, concerning the \textit{Architecture Coverage} factor,  these methods only provide partial interpretability. The capsule units in the capsule layer in Capsule Networks~\cite{capsul_net}, the interpretable filters in the last convolutional layer in Interpretable Convolutional Filter~\cite{interpretable_conv_filter} and Interpretable CNN Decision Tree~\cite{interpretable_decision_tree}, prototype layer in ProtoPNet~\cite{protopnet},  ProtoMIL~\cite{prtomil}, TesNet~\cite{tesnet}, and Deformable ProtoPNet~\cite{deformable_protopnet} as well as filters wrapped by the transformation module in CW~\cite{conceptwhitening} are the only units that become interpretable - the features encoded in the rest of the units that define the architecture (convolutional and fully-connected layers) are still opaque.

\subsubsection{Class-Agnostic}
\label{sec:int_b_dsg_ca}
One example of this category is the Case-based reasoning through prototypes as proposed in \cite{casebasedreasoning}. The method utilizes an autoencoder to reduce the dimensionality of the input and to learn useful
features for prediction.  This is followed by a prototype layer which introduces a set of prototypical vectors shared across classes. This is opposite to ProtoPNet~\cite{protopnet} which aims to learn representations which are very close or identical to an exemplar from the training set. 
We mentioned ProtoPNet as a \textit{Class-Specific} \textit{Interpretable-by-Design} method, since the method enforces the network to learn prototypical parts that are specific to each class by involving the distances to the specific prototypical vectors in classification task. In contrast, in \cite{casebasedreasoning} the distances to all the prototypical vectors will contribute to the probability prediction for each class. 
To make the class-specific prototypical vectors shareable among classes, ProtoPShare~\cite{protopshare} extends ProtoPNet~\cite{protopnet} by introducing a data-dependent similarity metric. This metric identifies similar learned prototypical vectors among classes, this is followed by a pruning step in order to reduce the number of prototypical vectors. More specifically, after the prototype layer is trained, the introduced data-dependency similarity computes the inverse of the distance between outputs of each pair prototypical vectors for all training images. Then, one of the similar/closest prototypical vectors is removed, and the weights of the remaining prototypical vectors are shared among both classes. Finally, the classifier part is fine-tuned.

Different from Case-based
reasoning~\cite{casebasedreasoning}, ProtoPNet~\cite{protopnet}, ProtoMIL~\cite{prtomil}, TesNet~\cite{tesnet}, Deformable ProtoPNet~\cite{deformable_protopnet}, and ProtoPShare~\cite{protopshare} that use a prototype layer, followed by a fully-connected layer as a classifier, ProtoTree~\cite{prtotree} utilizes a decision tree located after the final convolutional layer to learn a binary tree classifier of prototypical vectors shared among classes. Each node inside the tree is a trainable prototypical vector which is trained through a prototype learning procedure~\cite{protopnet}. Following this procedure, leave nodes learn class distributions. Hence, a path from the root to a leave represents the classification rule. Additionally, during the training phase each internal node (a prototypical vector) can generate a visualization for the training input sample which has the highest similarity (i.e., lowest \textit{$L^2$} distance) w.r.t. it. Therefore, ProtoTree is able to provide a hierarchical visualization of the decision-making process followed by the model.    

\textbf{Discussion.} Regarding the \textit{Annotation Dependency} factor, similar to the \textit{Class-Specific Interpretable-by-Design} methods, these methods are independent of any external annotations, and only rely on class labels.

Regarding the \textit{Feedback Modality} factor, ProtoTree~\cite{prtotree} and ProtoPShare~\cite{protopshare} are able to generate heatmap visualizations of the relevant units, i.e. learned prototypical vectors, as well as to extract image patches for the images which have the closest patch to one of the learned prototypical vectors in the tree. In contrast, Case-based reasoning~\cite{casebasedreasoning} does not provide a feedback visualization from the learned representations. It uses only the reconstructed input images as means of visualization. \figurename~\ref{fig:aic_hv} shows examples of the feedback modality in the ProtoTree and ProtoPShare methods.

Regarding the \textit{Feedback Semanticity} factor, it can be noted that similar to the \textit{Class-Specific Interpretable-by-Design} methods, Case-based reasoning~\cite{casebasedreasoning}, ProtoPShare~\cite{protopshare}, and ProtoTree~\cite{prtotree} do not 
guarantee any alignment w.r.t. semantic concepts.
The focus of these methods lies on justifying the prediction made by the model through the representations encoded in the prototype layer. 

Finally, concerning the \textit{Architecture Coverage} factor, similar to ProtoPNet~\cite{protopnet}, ProtoMIL~\cite{prtomil}, TesNet~\cite{tesnet}, and Deformable ProtoPNet~\cite{deformable_protopnet} the \textit{Class-Agnostic Interpretable-by-Design} methods namely, Case-based reasoning~\cite{casebasedreasoning}, ProtoPShare~\cite{protopshare}, and ProtoTree~\cite{prtotree} still suffer from the weakness of partial interpretability of the base model.

\begin{table*}[!t]
\renewcommand{\arraystretch}{1.3}
\centering
\caption{An integration of the
different aspects of the related works in terms of qualitative properties.  The column \textit{Feedback} Semanticity categorizes different feedback modalities into five categories: (1) Synthetic Images (SI), (2) Image Patches (IP), (3) Exemplar Images (EI), (4) Average Images Patches (AIP), and (5) Heatmap Visualization (HV).}
\label{table:qualitative_properties}

\begin{adjustbox}{width=\linewidth,center}
\begin{tabular}{lccccccc}
\hline
Methods & Interpretation Capability  & Task Specificity & Annotation Dependency & Feedback Modality &  Feedback Semanticity  &  Architecture Coverage & Explanation \\
 & Integration &  &  &  &  &  & Capability \\
\hline
\hline
Class Scoring Model 
\cite{class_scoring_model} & Post-Hoc & Class-Specific & Independent & SI & No & Partial & No \\
\cite{non_parametric_patch_prior} & Post-Hoc & Class-Specific & Independent & SI & No & Partial & No \\
\cite{mid_level_vp} & Post-Hoc & Class-Specific & Independent & IP & No & Partial & No \\
TCAV \cite{tcav} & Post-Hoc & Class-Specific & Independent & SI & No & All & No \\
VEBI \cite{vebi} & Post-Hoc & Class-Specific & Independent & AIP & No & All & Yes \\
ACE \cite{ace} & Post-Hoc & Class-Specific & Independent & IP & No & Partial& No  \\
Concept Attribution~\cite{concept_attribution} & Post-Hoc & Class-Specific & Independent & SI & No & All & No  \\
Critical Subnetworks~\cite{subnetworks} & Post-Hoc & Class-Specific & Independent & EI & No & All & Yes  \\
ICE~\cite{ice} & Post-Hoc & Class-Specific & Independent & IP & No & Partial & No  \\
PACE \cite{pace} & Post-Hoc & Class-Specific & Independent & IP & No & Partial& No  \\
\hline
\cite{object_detectors}& Post-Hoc & Class-Agnostic & Independent & IP & Yes & All & No 
\\
\cite{visual_att_indicator}& Post-Hoc & Class-Agnostic & Dependent & EI & Yes & All & No 
\\
Feature Inversion \cite{feature_inversion} & Post-Hoc & Class-Agnostic & Independent & SI & No & Partial & No \\
Network Inversion \cite{network_inversion} & Post-Hoc & Class-Agnostic & Independent & SI & No & Partial & No \\
Linear Probes~\cite{linear_probing} & Post-Hoc & Class-Agnostic & Independent & - & No & All & No 
\\
Network Dissection \cite{networkdissection}& Post-Hoc & Class-Agnostic & Dependent & IP & Yes & All & No
\\
Net2Vec \cite{net2vec}& Post-Hoc & Class-Agnostic & Dependent & IP & Yes & All & No  \\
Compositional Explanations~\cite{compositionalexplanations}& Post-Hoc & Class-Agnostic & Dependent & IP & Yes & Partial & No
\\
Selectivity Index ~\cite{selectivity_index} & Post-Hoc & Class-Agnostic & Independent & AIP & Yes & All & No 
\\
Topic-based interpretation \cite{topicmodel}& Post-Hoc & Class-Agnostic & Independent & IP & No & Partial & No
\\
Revers Linear Probing~\cite{revers_probing}
& Post-Hoc & Class-Agnostic & Dependent & EI & No & Partial & No  \\
~\cite{intervention}
& Post-Hoc & Class-Agnostic & Independent & SI & No & Partial & No  \\
\hline
Capsule Network\cite{capsul_net}& Interpretable-by-Design & Class-Specific & Independent & - & No & Partial & No  \\
Interpretable Convolutional Filter \cite{interpretable_conv_filter}& Interpretable-by-Design & Class-Specific & Independent & IP & Yes & Partial & No  \\
Interpretable Decision Tree~\cite{interpretable_decision_tree} & Interpretable-by-Design & Class-Specific & Independent & EI/IP & Yes & Partial & No  \\
ProtoPNet \cite{protopnet}& Interpretable-by-Design & Class-Specific & Independent & HV/IP & No & Partial & No  \\
CW \cite{conceptwhitening}& Interpretable-by-Design & Class-Specific & Independent & EI & No & Partial & No  \\
ProtoMIL \cite{prtomil}& Interpretable-by-Design & Class-Specific & Independent & HV & No & Partial & No  \\
TesNet \cite{tesnet}& Interpretable-by-Design & Class-Specific & Independent & HV/IP & No & Partial & No  \\
Deformable ProtoPNet~\cite{deformable_protopnet}& Interpretable-by-Design & Class-Specific & Independent & IP & No & Partial & No  \\
\hline
Case-based reasoning through prototype \cite{casebasedreasoning}& Interpretable-by-Design & Class-Agnostic & Independent & -- & No & Partial & No  \\
ProtoPShare \cite{protopshare}& Interpretable-by-Design & Class-Agnostic & Independent & HV/IP & No & Partial & No  \\
ProtoTree \cite{prtotree}& Interpretable-by-Design & Class-Agnostic & Independent & HV/IP & No & Partial & No  \\
\hline

\end{tabular}
\end{adjustbox}
\end{table*}


\section{Discussion}
\label{sec:discussion}

\begin{figure*}[t]
\centering
\includegraphics[width=0.8\linewidth]{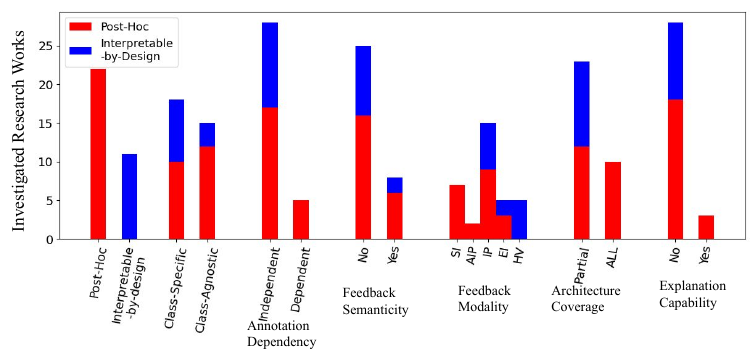}
\caption{Statistics of investigated visual model interpretation methods over different proposed qualitative factors.}
\label{fig:barchart}
\end{figure*}

In this section, we provide a discussion of the different type of interpretation methods over the defined factors (Table~\ref{table:qualitative_properties}).
Here, we further extend the focused category-specific discussions provided earlier in Section~\ref{sec:framework} by addressing the overarching trends observed in \figurename~\ref{fig:barchart}.
In this figure, the horizontal axis shows the qualitative properties while the vertical axis illustrates the number of works, covered in this study, which fall within a given property. The red and blue color indicate the \textit{Post-Hoc} and \textit{Interpretable-by-Design} categories, respectively. 
%


\subsection{Interpretation Capability Integration}
This factor describes interpretability capabilities are injected at design time, i.e. \textit{Interpretable-by-Design}, or in a
 \textit{Post-Hoc} manner. According to Table~\ref{table:qualitative_properties}, the history of methods following a \textit{Post-Hoc} approach is much older than that of \textit{Interpretable-by-Design} approaches. Also, as can be seen in \figurename~\ref{fig:barchart}, the majority of the reviewed interpretation methods have followed the \textit{Post-Hoc} approach. 
Possible reasons for such a trend could be the following. First, the interpretation of deep models was identified as a problem of interest following the seminal work of \cite{alexnet} and the remarkable results it obtained in the ImageNet ILSVRC'12 challenge.
As such, initial interpretation efforts were formulated for scenarios where the base models were already in place, i.e after the training phase. 
Second, the methodology followed by these methods do not need to touch the original model or its training procedure. 
Hence, these methods do not affect the inner-workings and performance of already pre-trained methods. This reduces the design complexity of the algorithms in this approach which makes it simpler than \textit{Interpretable-by-Design} approaches. 

The methods following the \textit{Post-Hoc} modality provide interpretations through a model approximation strategy. This raises questions regarding the fidelity or faithfulness of the provided interpretation feedback. More specifically, this issue is related to the level to which the provided interpretation feedback are faithful to the representations learned by the model.
Moreover, due to their characteristic of operating on top of an existing model, \textit{Post-Hoc} methods tend to require additional computations than their \textit{Interpretable-by-Design} counterparts. 
As consequence of these weaknesses, the \textit{Interpretable-by-Design} research line has recently become very active. The methods following this approach are able to reveal, to some level, the inference procedure followed by the model. However, the additional built-in modules and specific representation learning algorithms used by this type of methods increase their design complexity.

\subsection{Task specificity}
In this section, we discuss  
the \textit{Task Specificity} factor addressed by the model interpretation methods. As can be noted in  Figure~\ref{fig:barchart}, the number of proposed methods in each of the \textit{Class-Specific} and \textit{Class-Agnostic} categories  are roughly the same. However, it can be observed that the number of \textit{Class Agnostic Post-Hoc} methods is sligthly higher than that of \textit{Class-Specific Post-Hoc}. Moreover, we notice an opposite trend for the \textit{Interpretable-by-Design} category.
The reason for such trend lies on the methodology and goal  of each of these categories. The majority of \textit{Class-Specific Post-Hoc} methods, such as \cite{class_scoring_model,mid_level_vp,ace,subnetworks,ice,concept_attribution}, need to be run in separate stages considering a limited set of examples from one class at a time. This makes these methods computationally expensive. %
In contrast, \textit{Class-Agnostic Post-Hoc} methods follow algorithms that do not require this splited processing of the data. 

In the \textit{Interpretable-by-Design} side, \textit{Class-Agnostic} methods,  have the inherent weakness of not being able to directly link the provided interpretation feedback with the classes of interest. 
To cope with such limitation, recently, the \textit{Interpretable-by-Design} research line has been more headed towards learning class-relevant interpretable representations. Consequently, this group of methods provide better insights on the learned representations and their relationship w.r.t. the classes of interest. This characteristic is important for fine-grained classification problems where the subtle differences among the categories are of interest.


\subsection{Annotation dependency and feedback semanticity}
\label{sec:annotation_dependency_feedback_semanticity}
This section investigates the capability of the interpretation methods in providing semantic feedback as well as their dependency on annotations.

According to \figurename~\ref{fig:barchart}, the \textit{Feedback Semanticity} and \textit{Annotation Dependency} factors follow, more or less, similar trends. There are several observations that can be made from here.

First, while majority of the visual model interpretation methods are not able to provide semantic feedback, they are also independent of any additional external annotations.

Second, there is a small group of methods that provide semantic feedback. To do so, these methods use additional annotations of different forms. For example, Network Dissection~\cite{networkdissection}, Net2Vec~\cite{net2vec}, and Compositional Explanation~\cite{compositionalexplanations} utilize pixel-level annotations for pre-defined object classes in the dataset. \cite{visual_att_indicator} and Reversed Linear Probing~\cite{revers_probing}, on the other hand, rely on image-level attribute annotations.

Third, all the \textit{Interpretable-by-Design} methods are independent of any external annotations. This limits them from providing any semantically-meaningful feedback.
Considering this, extending existing methods to provide semantic feedback and their quantitative evaluation w.r.t. those provided by \textit{Post-Hoc} methods can be another research problem in the field. Also, the most recently proposed methods in the \textit{Post-Hoc} category do not explore systemically the semantics of their provided interpretation. Hence, developing the methods in this category to provide semantic feedback is still an open problem.


\subsection{Feedback Modality}
\label{sec:feedback_modality}
\begin{figure*}[t]
\centering
\includegraphics[width=0.8\linewidth]{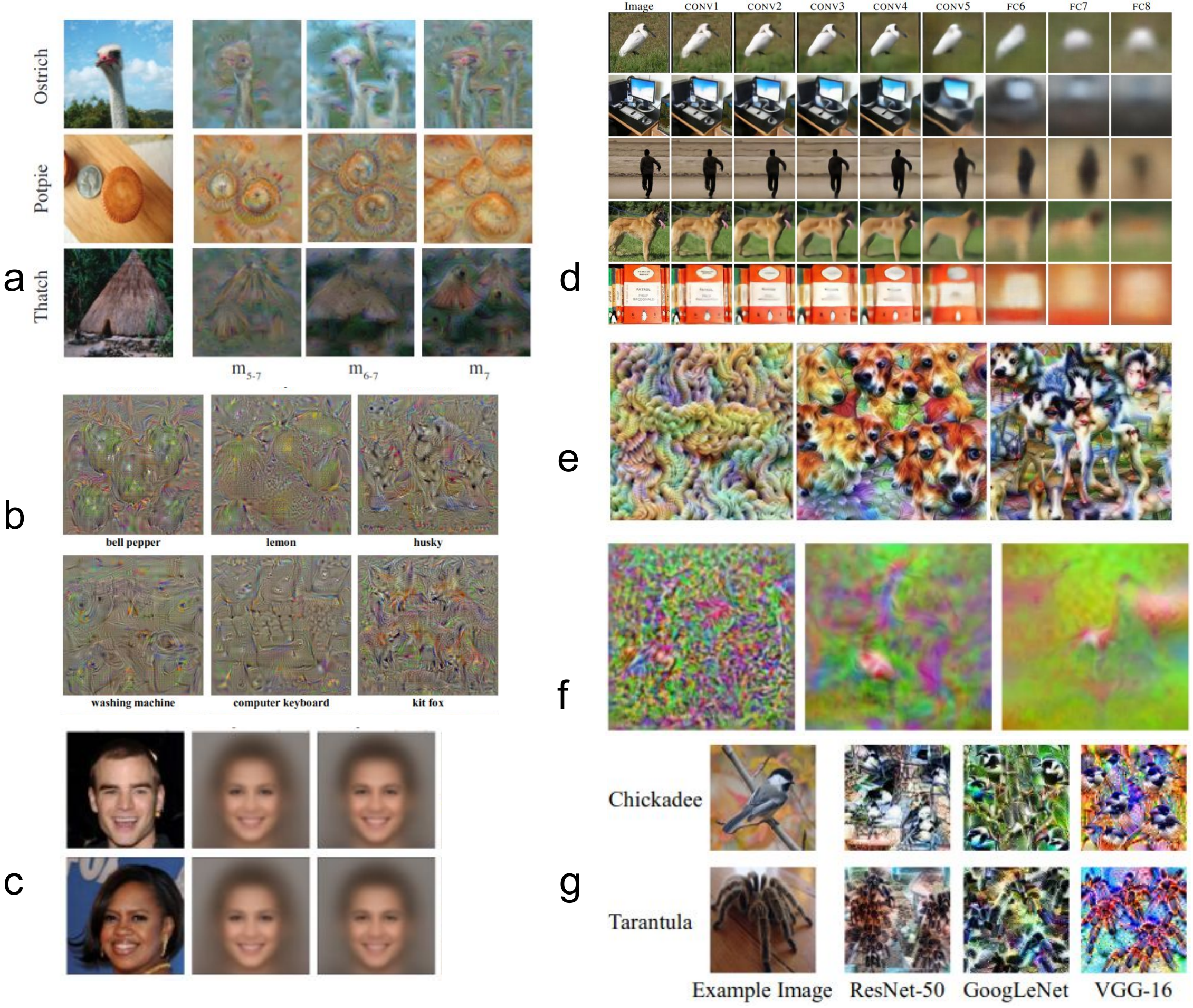}
   \caption{ Example of Synthetic Images (SI) feedback modality illustrated in a)~\cite{non_parametric_patch_prior}, b) Class Scoring Model~\cite{class_scoring_model}, c)~\cite{intervention}, d) Network Inversion~\cite{network_inversion}, e) TCAV~\cite{tcav}, f) Feature Inversion~\cite{feature_inversion}, g) Concept Attribution~\cite{concept_attribution}.}
\label{fig:si}
\end{figure*}

\begin{figure*}[t]
\centering
\includegraphics[width=0.8\linewidth]{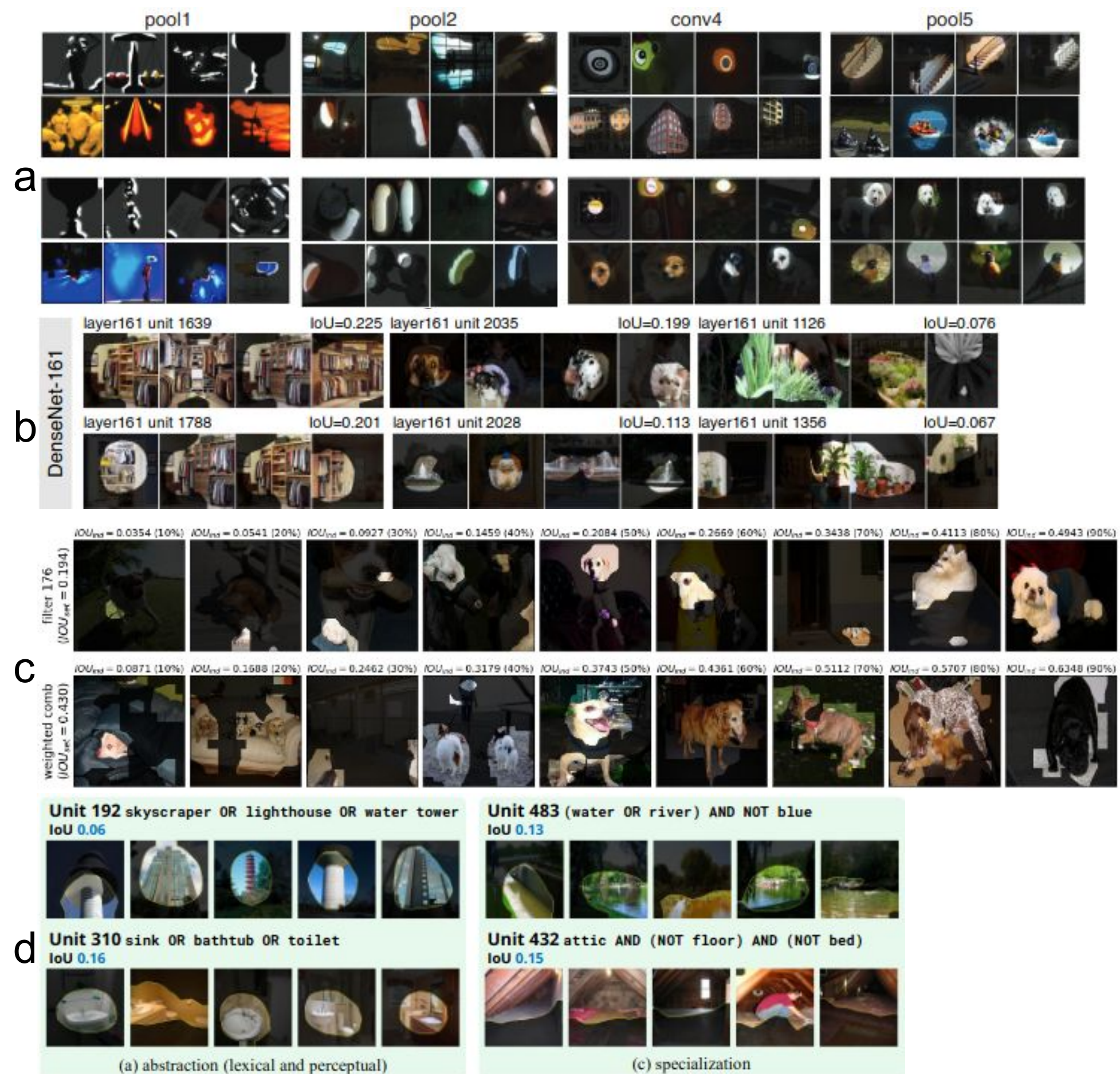}
\caption{Example of Images Patches feedback modality illustrated in a)~\cite{object_detectors}, b) Network Dissection~\cite{networkdissection}, c) Net2Vec~\cite{net2vec}, and d) Compositional Explanation~\cite{compositionalexplanations}.}
\label{fig:ip_1}
\end{figure*}

\begin{figure*}[t]
\centering
\includegraphics[width=0.8\linewidth]{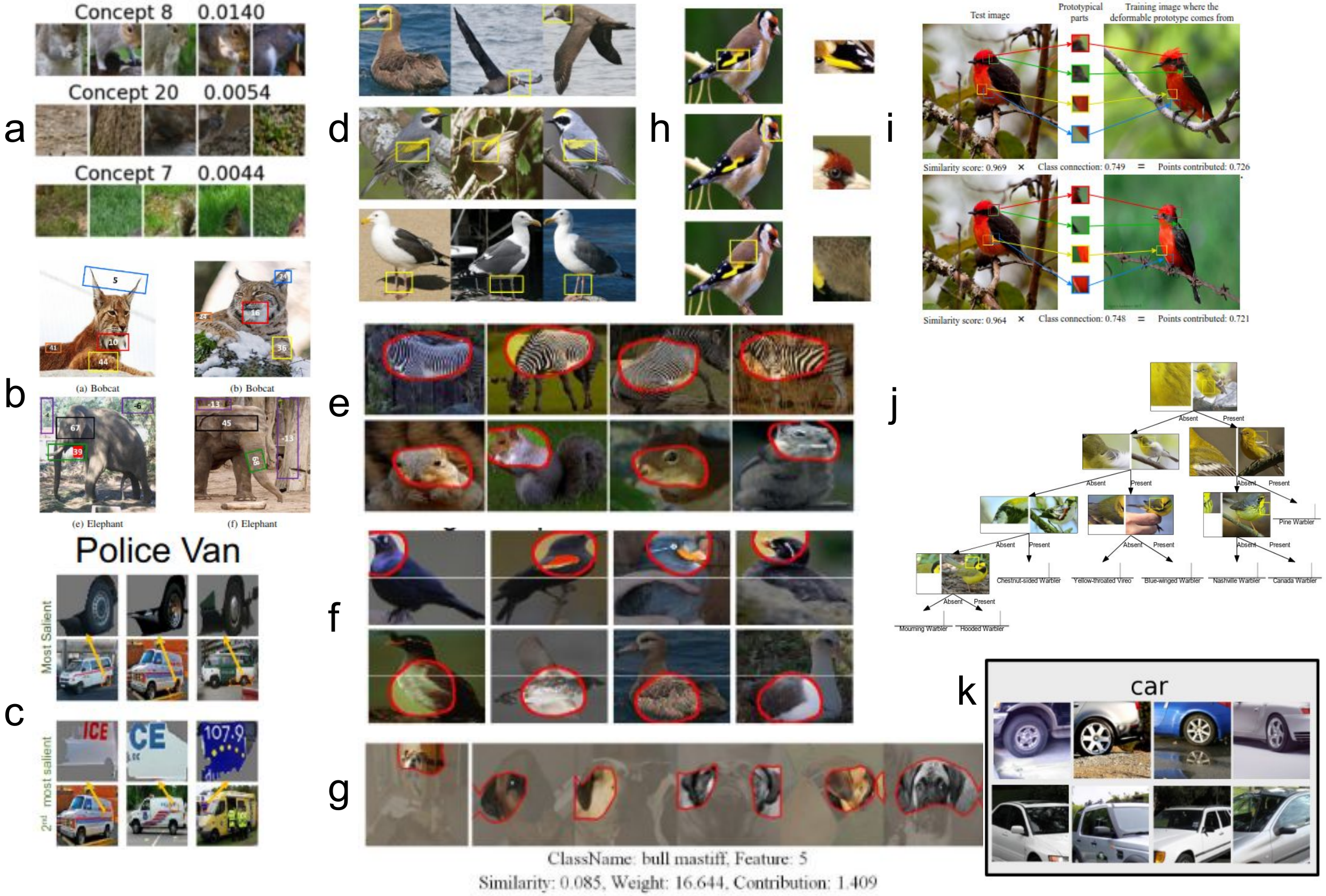}
   \caption{Example of Image Patches feedback modality illustrated in a) Topic-based interpretation~\cite{topicmodel}, b) PACE~\cite{pace}, c) ACE~\cite{ace}, d) ProtoPNet~\cite{protopnet}, e) Interpretable Convolutional Filter~\cite{interpretable_conv_filter}, f) Interpretable CNN Decision Tree~\cite{interpretable_decision_tree}, g) ICE~\cite{ice}, h) TesNet~\cite{tesnet}, i) Deformable ProtoPNet~\cite{deformable_protopnet}, j) ProtoTree~\cite{prtotree}, and k)~\cite{mid_level_vp}.
   }
\label{fig:ip_2}
\end{figure*}

\begin{figure*}[t]
\centering
\includegraphics[width=0.8\linewidth]{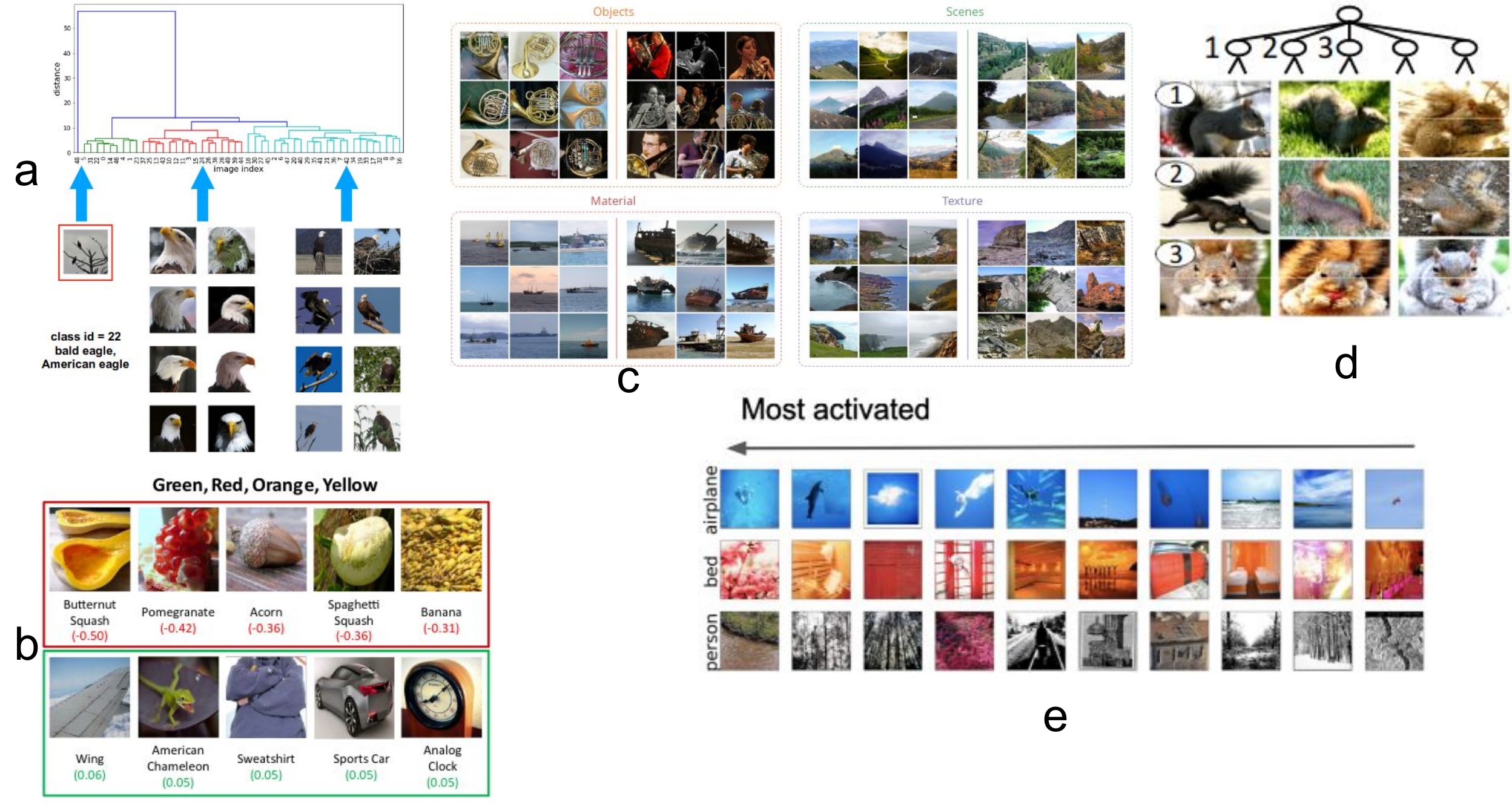}
\caption{Example of Images feedback modality illustrated in a) Critical Subnetwork~\cite{subnetworks}, b)~\cite{visual_att_indicator}, c) Reversed linear Probing~\cite{revers_probing}, d) Interpretable CNN Decision Tree~\cite{interpretable_decision_tree}, and e) Concept Whithening~\cite{conceptwhitening}. }
\label{fig:i}
\end{figure*}

\begin{figure*}[t]
\centering
\includegraphics[width=0.8\linewidth]{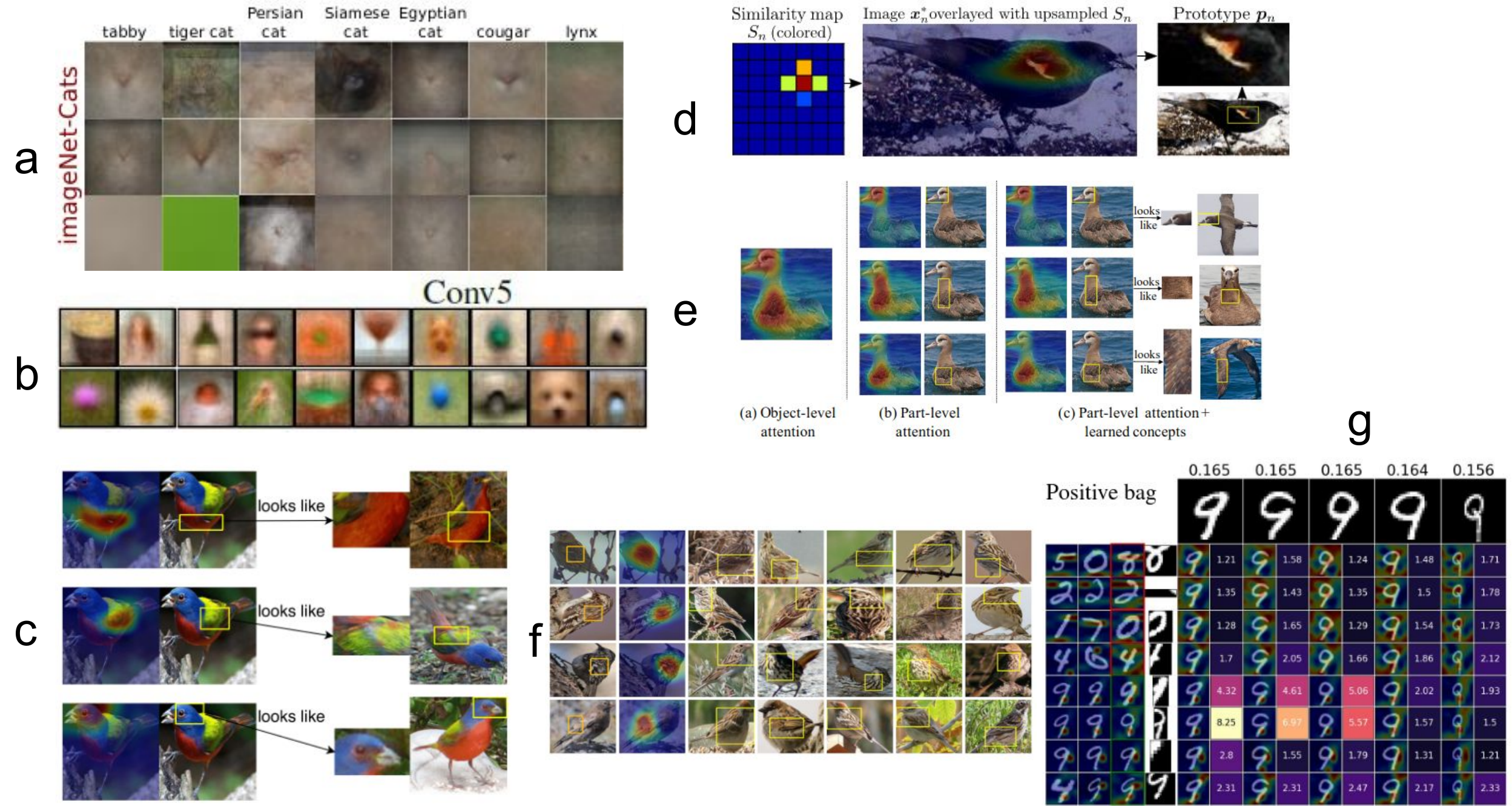}
   \caption{a and b show example of Average of Image Patches (AIP) feedback modality illustrated in VEBI~\cite{vebi} and Selectivity Index~\cite{selectivity_index}. The other images show the example of Heatmap Visualization (HV) feedback modality presented in c) ProtoPNet~\cite{protopnet}, d) ProtoTree~\cite{prtotree}, e) TesNet~\cite{tesnet}, f) ProtoPShare~\cite{protopshare}, and g) ProtoMIL~\cite{prtomil}.}
\label{fig:aic_hv}
\end{figure*}

This factor describes the form of the feedback provided by a given  interpretation method. Based on the discussion conducted in Section~\ref{sec:framework}, we can categorize different feedback modalities into five categories: (1) Synthetic Images (SI) (\figurename~\ref{fig:si}), (2) Image Patches (IP) (\figurename~\ref{fig:ip_1} and \ref{fig:ip_2}), (3) Exemplar Images (EI) (\figurename~\ref{fig:i}), (4) Average Images Patches (AIP) (\figurename~\ref{fig:aic_hv}.a and b), and (5) Heatmap Visualization (HV) (\figurename~\ref{fig:aic_hv}.c-f). Statistics on the  use of these modalities is presented in \figurename~\ref{fig:barchart}. As can be seen, Exemplar Images (EI) and Images Patches (IP) are the most used modality to visualize the interpretation feedback among the visual model interpretation methods. 

Considering the \textit{Post-Hoc} category, none of the methods produces heatmap visualizations as part of their provided interpretation feedback.  Also, while generating Image Patches (IP) is the most used feedback modality, Average Image Patches (AIP) is the second less common feedback modality used by the methods.
This suggests that the identified/learned relevant features might not be analyzed with an appropriate level of depth. This observations arises from the difference between these two types of feedback modality. While IP-based feedback usually highlights the patches with the highest response, the AIP counterpart stress the consistency among the patches.  

Regarding the \textit{Interpretable-by-Design} category, it is noticeable that none of the methods uses Synthetic Images (SI) or Average of Images Patches (AIP) as their feedback modality. 
Using these modalities to shed light on the features learned
by the \textit{Interpretable-by-Design} methods can provide a deeper intuition on them. As can be seen in \figurename~\ref{fig:barchart}, the majority of the \textit{Interpretable-by-Design} methods can generate Image Patches. Therefore, visualizing the average of images patches, for example in prototype-based methods, can reveal whether the learned prototypical vectors consistently align around a specific visual pattern. 
Moreover, utilizing the activation maximization technique~\cite{nmf} at test time to generate synthetic images whose feature maps have closest distance (highest similarity) to a given learned prototypical vector can provide a general conceptual intuition on the encoded visual pattern. 
Furthermore, in the case that a method is not able to generate image patches nor a heatmap visualization, e.g. Concept Whitening~\cite{conceptwhitening}, the activation maximization technique~\cite{nmf} can be an alternative approach. %
Concept Whitening~\cite{conceptwhitening} tries to separate the direction of different encoded features in the latent space. As mentioned, the method illustrates only exemplar images for each direction in the latent space. Therefore, there is no guarantee that a clear and consistent visual pattern is illustrated in the exemplar images provided as interpretation feedback. 
Therefore, generating synthetic images for each direction can provide a better intuition on the encoded features in each direction of the latent space. This would ease the qualitative assessment on whether the latent space directions have encoded distinctive visual patterns.
Similarly, the activation maximization technique~\cite{nmf} can be applied on the trained capsules layers in Capsule Networks~\cite{capsul_net}.
Since the Capsule Networks aim at learning class-specific capsules using routing-by-agreement mechanisms, one can generate a synthetic image whose internal representation maximizes the output probability vectors of capsules relevant to a given class. Then, the relationship among the learned features can be evaluated from the synthetic images.

\subsection{Explanation capability}
This section discusses whether the insights extracted via interpretation methods could be further exploited for model explanation purposes. More specifically, to justify the prediction made by the model for a specific input. 
While this capability is not a necessary requirement for model interpretation methods, its existence would further extend the value and applicability of the interpretation method that possess it. This explanation capability can be in a form of a visualization, e.g. as proposed in \cite{vebi,subnetworks}. We refer to this visualization modality as explanation feedback based on the relevant units extracted by the interpretation method for each input image respective to the decision made by the model. This is different from interpretation visualization in the \textit{Feedback Modality} factor where the interpretation methods provide interpretation feedback, i.e., visual feedback of the identified/learned relevant features, regardless of the predictions made by the model.

The statistics of the \textit{Explanation Capability} factor are presented in \figurename~\ref{fig:barchart}. From here we can draw the following observations and analysis. 

It is noticeable that only two of the \textit{Post-Hoc} interpretation methods, namely VEBI~\cite{vebi} and Critical Subnetwork~\cite{subnetworks}, are able to explain the predictions of the model based on the insights extracted by the interpretation method. 

VEBI generates visual explanations by 
computing visual explanations for the relevant features (identified by the interpretation procedure) related to the class label predicted for a given input.

Critical Subnetwork follows a Grad-Cam-like method~\cite{gradcam} to generate visual explanations. 

Given an extracted \textit{Class-Specific} subnetwork, Critical Subnetwork computes the gradient of the output of the predicted class only w.r.t the convolutional filters identified as class-relevant units. Then, the gradient maps pertaining to each class-relevant convolutional filters are considered for visualization. 

Regarding \textit{Interpretable-by-Design} methods, it can be seen that none of them generate visual explanations. Therefore, enhancing these methods with this capability could be point of action for future work. This capability, specifically for prototype-based methods, enables the discovery of critical learned \textit{Class-Specific} or \textit{Class-Agnostic} 
prototypical vectors which can be used for justifying the decision-making process of a model. For example, gradients of the model output w.r.t class-specific prototypical vectors can be computed to produce a saliency map that highlight important region(s) in the provided input.

 
\subsection{Architecture coverage}
In this section, we focus the  discussion on the portion of the architecture from where insights are extracted by interpretation methods. 
Studies have revealed that different features are encoded in units/neurons located at different levels of the architecture. 
 
Hence, the produced interpretation feedback should ideally be in accordance with features encoded in all the layers or parts of the architecture of the model being interpreted. The statistics illustrated in \figurename~\ref{fig:barchart} show that only a small set of interpretation methods, which are \textit{Post-Hoc}, provide interpretation by considering all the convolutional layers of a given architecture. 
Table~\ref{table:qualitative_properties} shows that recent \textit{Class-Specific} and \textit{Class-Agnostic} \textit{Post-Hoc} methods have a partial coverage of the architecture of the base model. Here a common practice is to focus on the last convolutional layer of a given architecture.  
Hence, the produced interpretation feedback is limited to a small part of the model. Furthermore, none of the \textit{Interpretable-by-Design} methods consider the representations encoded in all the layers. Here again, a common practice is to focus on the representations encoded at the last convolutional layer. This, in turn,  reduces the level of insights they are capable of producing in their interpretation feedback. 
In this regard, extending \textit{Interpretable-by-Design} methods, for example investigating the possibility of applying the prototype layers in all the parts of the architecture, can be a potential research direction in this field.


\subsection{Evaluation protocol}
In this section, we discuss the evaluation protocols followed by the discussed methods to assess the performance of the produced interpretation feedback. These protocols cover both  qualitative and quantitative evaluation.  

According to our study, we have observed that, with the exception of Network Dissection~\cite{networkdissection} and its extensions namely; Net2Vec~\cite{net2vec} and Compositional Explanation~\cite{compositionalexplanations}, the other visual interpretation methods only provide qualitative examples as interpretation feedback (Section~\ref{sec:feedback_modality}).  

In the cases where a quantitative evaluation is conducted, the methods follow different approaches and goals to quantify the produced interpretation feedback. 

User-based evaluation is one of the approaches conducted in \cite{object_detectors}, ACE~\cite{ace}, and PACE~\cite{pace}. These user-studies are based on questionnaires which include questions about the occurrence of a semantic concept in the visual feedback provided by a given interpretation method. 
Then, a group of users are asked to assign scores to the presented visualizations. These scores aim to indicate the level of agreement of the users w.r.t. the predefined concepts. Finally, the 
interpretation capability of a method is quantified by aggregating the collected scores.

A group of methods, namely Class Scoring Model~\cite{class_scoring_model}, Feature Inversion~\cite{feature_inversion}, and Network Inversion~\cite{network_inversion} report the representation reconstruction error as a quantitative metric to measure the performance of the produced interpretation feedback. 

In other cases where the aim is to quantify the semanticity of the internal representation, the alignment between annotation masks and internal activations is measured. This is specifically performed in Network Dissection~\cite{networkdissection}, Net2Vec~\cite{net2vec},  Compositional Explanation~\cite{compositionalexplanations}, and Interpretable Convolutional Filters~\cite{interpretable_conv_filter}. 

In some methods, such as \cite{intervention} and TCAV~\cite{tcav}, the cosine similarity between the representation internally-encoded by the base model and the representation learned/identified by the interpretation methods is measured.
In other cases, namely Linear Probing~\cite{linear_probing}, classifiers are trained on the internal representations of each layer, separately. Similarly, in Reversed Linear Probing~\cite{revers_probing} and CW~\cite{conceptwhitening} classifiers are trained on the representations defined by the learned relevant components to measure separability of the learned representations. Then, the classification accuracy of 
these two groups of classifiers are compared for as part of the evaluation.

Worth noting is that these classifiers are different from the base model being interpreted.

In various cases~\cite{visual_att_indicator}, \cite{vebi}, \cite{subnetworks} a quantitative evaluation is conducted to investigate the relevance of the units identified by the interpretation methods.
However, such evaluation is designed with different approaches closely related to the proposed methodology. 
For instance, \cite{visual_att_indicator} and VEBI~\cite{vebi} apply a neuron-perturbation approach where the identified relevant units are systematically  occluded by zeroing their output. Then, examples are pushed through the perturbed model and the changes in classification accuracy are tracked. The assumption here, is that the occlusion of the relevant units should lead to significant drops in classification performance.
Critical Subnetwork~\cite{subnetworks} and Interpretable CNN Decision Tree~\cite{interpretable_decision_tree} compare the output accuracy between the original model and the extracted one, i.e. the extracted subnetwork or constructed decision tree, respectively. 
In this regard, the topic-based interpretation method~\cite{topicmodel}  proposes the ConceptShAP metric, adapted from Shapley Values~\cite{shapley}, to measure the completeness score of the learned topics for a  model prediction. TCAV~\cite{tcav} measures the sensitivity of the output logits of the model by considering the difference between the output logit for the original activations and the activations aggregated by the obtained CAV. In this line, ACE~\cite{ace}, Concept Attribution~\cite{concept_attribution}, and CW~\cite{conceptwhitening} have adapted TCAV to assign an importance score to their produced interpretation feedback w.r.t. the model accuracy. 

As can be seen, there is a clear diversity among the evaluation protocols followed by previous efforts. Furthermore, the followed protocols are tailored to the inner-workings of each interpretation method. This makes a uniform quantitative  comparison among existing model interpretation methods problematic. 
To address this problem, \cite{uniform_eval_protocol} has recently proposed an evaluation protocol that aims at the quantitative comparison of visual model interpretation methods. More specifically, the proposed method measures the alignment between heatmaps produced from relevant units identified by model interpretation methods and additional annotated semantic elements. These annotations have different levels of semanticity, e.g. objects, parts, colors, textures, and come from the same dataset used to train the model being interpreted.


\section{Related Work}
\label{sec:related_work}

\subsection{Model Explanation methods}
Model explanation methods aim to justify the prediction made by a model for a specific input~\cite{expltaxonomy}, \cite{beyond_saliency}, \cite{grounding_vis_expln}. Up to now, this research line has been significantly explored, thus introducing a wide terminology and a variety of  approaches~\cite{interpretation_survey_2}.  
Recently, \cite{expl_meta_survey} conducted a meta-study on the latest 20 most cited taxonomies/surveys, covering up to 50 model explanation methods, in order to highlight the essential aspects of the state-of-the-art in model explanation. Compared to the model explanation research line, a systematic study of model interpretation methods has remained non-existent. 

Different from the above mentioned surveys, \cite{stop_explaining} discusses issues such as faithfulness raising in model explanation methods in high stakes applications. 
Accordingly, it encourages policy makers towards using interpretable machine learning models instead of following post-hoc explanation procedures. Furthermore, the work investigates some challenges in the design of interpretable machine learning. These challenges  include architecture design, optimization algorithm construction, and scarcity of domain experts for the analysis of  the feedback provided by interpretable machine learning models in high-stakes applications.

Here, we have provided a framework to classify visual model interpretation methods according to the defined axes/factors. The proposed framework reveals the 
strengths and drawbacks of current model interpretation efforts. It also sheds light on possible research gaps in this line of research that can be explored further.

\subsection{Model Interpretation Methods}
Model interpretation methods aim to analyze the internal representations learned by the model. 
One of the early related taxonomies in this research line was published in 2018 by \cite{interpretation_survey}, a review on several research directions in the area of visual interpretability of CNN models. These directions cover some works in the following areas: (1) visualizing the internal representations, (2) diagnosing representation flaws in CNNs, (3) disentangling internal representations into graphical models, (4) developing the architecture of CNNs to include built-in modules, such as R-CNN and Capsule Networks, to process different patterns in the internal representations. 
Although this was the first time that some of these research lines were reviewed, there are some gaps in that study.  
First, it is important to note that while it provides insights into the works proposed along each direction, it did not specify which were the factors to be studied over methods nor accomplished a comparison/discussion between works w.r.t their qualitative properties.
Second, it considers model explanation methods as part of the research line of visual model interpretation. As stated earlier explanation methods have a different goal than their interpretation counterparts. 
Third, it covers works along the direction of diagnosing representation flaws in CNNs, such as bias detection in the learned representations. While a very important problem, detecting bias can be considered as a task that can be assisted by model interpretation methods but not a goal of interpretation methods per se. 

Another survey in this research line, \cite{interpretation_survey_2} has proposed a taxonomy to categorize methods based on three axes namely (1) relation between interpretation methods and the model being interpreted, (2) type of explanations provided by the covered methods, and (3) local vs. global interpretability. Most recently, \cite{interpretation_survey_3} put forward a new survey on model interpretation to classify methods based on three axes: (1) representations of interpretations, e.g., the input feature importance or the influece of the training samples, (2) type of the base model that the interpretation method is used for, e.g., differentiable models, GANs, and NLP models, and (3) relation between interpretation methods and the model being interpreted (similar to \cite{interpretation_survey_2}).

Although these works have provided a comprehensive survey of existing methods,  they suffer from some weaknesses as well. 
First, from the point of terminology, these works use frequently the terms of \textit{explanation} and \textit{interpretation} interchangeably, which lends itself to confusion. 
Second, they cover a wide variety of explanation methods.
Moreover, since \cite{interpretation_survey_3} considers a variety of deep models, it provides a few examples of interpretation methods for each type of deep model. Hence, there are plenty of visual model interpretation methods that had not been covered by these surveys. Furthermore, it considers the methods using attention mechanisms for explaining deep models as interpretation methods. However, as \cite{attention_explanability} has showed, purely focusing on  attention mechanisms might not be sufficient for this task.

In recent years, research along the visual model interpretation   line has gained significant momentun. This in turn led to works introducing new methodologies in the field. 
Hence, in this work, we have provided a framework with specific factors which could serve as axes for positioning existing and future methods. More important, we  provide a clear definition for the model interpretation task and covered in our study the methods compatible with it. 
We also provide inter-category and intra-category
discussion in order to give deeper insights on the interpretation capability of existing methods. 
We expect this could serve as a foundation for the model interpretation research line and could help reveal active and passive areas of research. 
In these regard, we stress some of the research gaps in each category. 


\section{Conclusion}
\label{sec:concolusion}
In recent years, research on methods for analyzing the representations internally encoded by Convolutional Neural Networks (CNNs) has been increased significantly. 
This increased interest, next to the continuously growing literature on the model explanation task, has produced the side effect of an increasing number of works with confusing use of terminology, e.g. "interpretation" vs "explanation". This not only leads to ambiguity and confusion but also hinders the identification of unexplored research areas/problems in the field. 
Here, we aim at making a clear distinction between these two tasks, i.e. model explanation and model interpretation, and 
conducted a detailed study of works addressing the latter.

A key contribution of our study of interpretation methods is the proposed framework, defined by six qualitative factors, that can serve for the categorization of current and future interpretation methods.
Accordingly, this document complements the description of existing interpretation methods with their positioning based on the proposed factors.

Following the proposed framework, we  highlighted several directions (e.g. reduced feedback semanticity, partial model coverage, etc. ) where research on model interpretation has received low attention. 
At the same time, we draw several pointers that could be followed to address such weaker directions. Finally, we discussed the evaluation protocols followed by each of the covered methods.

\ifCLASSOPTIONcompsoc

  \section*{Acknowledgments}
  This work is supported by the UAntwerp BOF DOCPRO4-NZ Project (id 41612) "Multimodal Relational Interpretation for Deep Models".
\else

  \section*{Acknowledgment}
\fi

\ifCLASSOPTIONcaptionsoff
  \newpage
\fi

\bibliographystyle{IEEEtran}
\bibliography{references}

\end{document}